\documentclass[11pt]{article}

\usepackage[preprint]{acl}

\usepackage{times}
\usepackage{latexsym}
\usepackage{booktabs}
\usepackage{siunitx}

\usepackage[T1]{fontenc}
\usepackage[utf8]{inputenc}

\usepackage{microtype}

\usepackage{inconsolata}

\usepackage{graphicx}

\title{MedSynth: Realistic, Synthetic Medical Dialogue-Note Pairs}

\author{ \textbf{Ahmad Rezaie Mianroodi}\textsuperscript{1,2} \quad \textbf{Amirali Rezaie}\textsuperscript{3} \quad \textbf{Niko Grisel Todorov}\textsuperscript{4} \quad \textbf{Nadine A. Friedrich}\textsuperscript{5} \\ \textbf{Maria P. Mogollon}\textsuperscript{5} \quad \textbf{Alexander Hernandez-Tirado}\textsuperscript{5} \quad \textbf{Guillermo Lopez Garcia}\textsuperscript{5} \\ \textbf{Cyril Rakovski}\textsuperscript{4} \quad \textbf{Frank Rudzicz}\textsuperscript{1,2} \\[0.6em] \textsuperscript{1}Dalhousie University, Canada \quad \textsuperscript{2}Vector Institute, Canada \quad \textsuperscript{3}Shahrood University of Technology, Iran \\ \textsuperscript{4}Chapman University, USA \quad \textsuperscript{5}Cedars-Sinai Medical Center, USA \\[0.4em] \small \texttt{ahmad.rm@dal.ca} }

\begin{document}
\maketitle
\begin{abstract}
Physicians spend significant time documenting clinical encounters, a burden that contributes to professional burnout. To address this, robust automation tools for medical documentation are crucial. We introduce MedSynth -- a novel dataset of synthetic medical dialogues and notes designed to advance the Dialogue-to-Note (Dial-2-Note) and Note-to-Dialogue (Note-2-Dial)  tasks. Informed by an analysis of disease distributions, this dataset includes over 10,000 dialogue-note pairs covering over 2000 ICD-10 codes. We demonstrate that our dataset substantially enhances the performance of models in generating both medical notes from dialogues, and dialogues from medical notes. The dataset provides a valuable resource in a field where open-access, privacy-compliant, and diverse training data are scarce. Code and sample of the data are available at \url{https://anonymous.4open.science/r/MedSynth/}.
\end{abstract}

\section{Introduction}

Electronic health records (EHRs) are widely used tools meant to streamline patient care \cite{melton2021electronic}. However, despite their benefits, EHRs also introduce a significant documentation burden on healthcare providers. Physicians report spending between 52 to 102 minutes daily on clinical note-taking from patient encounters, a task that is both time-consuming and detracting from direct patient care \cite{hripcsak2011use}. This extensive documentation requirement contributes significantly to physician burnout, highlighting an urgent need for solutions that can mitigate these demands while preserving the quality and integrity of patient records.

Automated tools for medical documentation have emerged as promising solutions to reduce this burden. An important consideration in developing these tools is ensuring that generated medical notes adhere to standard structures to ensure continuity of care and facilitate effective communication among healthcare professionals. Among these standards, the SOAP (Subjective, Objective, Assessment, and Plan) format is widely used in primary care settings \cite{mathioudakis2016keep,podder2022soap}. However, developing and validating automated documentation tools is often hampered by the lack of large, open-access datasets that are both comprehensive and privacy-compliant \cite{yim2023aci}. Existing datasets are frequently limited in scope, cover only a few medical conditions, lack adherence to standard medical formats, or are generally unavailable due to privacy concerns, significantly restricting their utility and hampering progress \cite{wang2023notechat,yim2023aci}.

Data augmentation offers an effective approach to bridge this gap. While synthetic data may not capture all nuances of real patient data, it can still provide valuable diversity and volume, enhancing ML training processes without compromising patient privacy.
In this paper, we address the limitations of existing datasets by providing a novel synthetic dataset for data augmentation. 
Our primary contributions are:

\begin{itemize}\setlength\itemsep{0em}
\item \textbf{A large synthetic medical dialogue-note dataset:} We provide a comprehensive, privacy-compliant dataset of medical dialogues paired with clinical notes encompassing a wide array of medical conditions over more than 2000 different ICD-10\footnote{10$^{th}$ revision of the International Classification of Diseases (ICD) by the World Health Organization (WHO).} codes and over 10,000 dialogue-note pairs.  The notes in our dataset follow the SOAP structure (Subjective, Objective, Assessment, Plan), which is one of the most common formats of medical notes  \cite{Podder2023SOAP}. The diversity of  ICD-10 codes and the use of SOAP structure mimic the use case of primary care, which is an underrepresented specialty in AI research but generalizes to other specialties as well. 
\item \textbf{Novel data generation pipeline:} We introduce a novel pipeline for generating synthetic medical dialogue-note pairs, which is extensible to other domains. 
\item \textbf{SotA Dial-2-Note and Note-2Dial  models:} We release models fine-tuned on our dataset and achieve SotA performance in generating medical notes from doctor-patient dialogues, and vice versa. These models can be used to further develop tools to automate documentation.
\end{itemize}

\section{Related Work}

Several datasets are commonly used in medical dialogue-to-note summarization. 3M Health \cite{zhang2021leveraging}, Abridge \cite{krishna2020generating}, Augmedix \cite{yim2021towards}, emr.ai \cite{finley2018dictations}, and Nuance \cite{enarvi2020generating} are important datasets for developing and testing models but their lack of public availability restricts their utility for reproducible research.

In contrast, open-source datasets such as MTS-dialogue \cite{abacha2023empirical}, PriMock57 \cite{korfiatis2022primock57}, Aci-Bench \cite{yim2023aci}, and NoteChat \cite{wang2023notechat} represent a meaningful advancement in improving the accessibility of medical dialogue data. Notably, some of these datasets, such as Aci-Bench, include synthetic dialogues created through human role-playing while others, like NoteChat \cite{wang2023notechat}, use large language models (LLMs) to generate dialogues. 

However, these open datasets still have various limitations. They often concentrate on specific diseases --- lacking a comprehensive range of medical conditions common in primary care -- or they do not follow standard medical note structures. Table~\ref{tab:dataset_comparison} shows a comparison of the available datasets.

\begin{table*}[!ht]
\small
  \centering

  % {\small
  \begin{tabular}{>{\raggedright\arraybackslash}p{0.15\textwidth} >{\raggedright\arraybackslash}p{0.25\textwidth} >{\raggedright\arraybackslash}p{0.15\textwidth} >{\raggedright\arraybackslash}p{0.1\textwidth} >{\raggedright\arraybackslash}p{0.08\textwidth} >{\raggedright\arraybackslash}r >{\raggedright\arraybackslash}p{0.05\textwidth}}
    \toprule
    \textbf{Dataset} & \textbf{Description} & \textbf{Tasks} & \textbf{Src-len (tok/turns)} & \textbf{Target-len (tok/sent)} & \textbf{Size} & \textbf{Open} \\
    \midrule
    3M Health \cite{zhang2021leveraging} & Dialogue-note pairs: notes are created using dialogues & Dial-2-Note, Note-2-Dial & -/- & -/- (HPI only) & 1342 & No \\
    \midrule
    Abridge \cite{krishna2020generating} & Dialogue-note pairs: notes are created using dialogues & Dial-2-Note, Note-2-Dial, ASR & 1500/- & -/27 & 6862 & No \\
    \midrule
    Augmedix \cite{yim2021towards} & Real clinical dialogue-note pairs & Clinical Dial-2-Note sentence alignment, Dial-2-Note & -/175 & -/47 & 500 & No \\
    \midrule
    EMR.AI \cite{finley2018dictations} & Real clinical dialogue-note pairs & Clinical form filling & 616/1 & 550/- & 9875 & No \\
    \midrule
    Nuance \cite{enarvi2020generating} & Real clinical dialogue-note pairs & Dial-2-Note, Note-2-Dial & 972/- & 452/- & 802k & No \\
    \midrule
    MTS-dialogue \cite{abacha2023empirical} & Dialogue-note snippets: dialogues are created using clinical note sections & Dial-2-Note, Note-2-Dial & 142/9 & 48/3 & 1701 & Yes \\
    \midrule
    PriMock57 \cite{korfiatis2022primock57} & Dialogue-note pairs: role-played by human & Dial-2-Note, Note-2-Dial, ASR & 1489/97 & 161/23 & 57 & Yes \\
    \midrule
    Aci-Bench \cite{yim2023aci} & Dialogue-note pairs: role-played by human & Dial-2-Note, Note-2-Dial & 1,302/55 & 490/37 & 207 & Yes \\
    \midrule
    NoteChat \cite{wang2023notechat} & Dialogue-note pairs: dialogue role-played by LLMs based on patient summaries & Dial-2-Note, Note-2-Dial & 561/27 & 301/14 & 207k & Yes \\
    \midrule
    \textbf{MedSynth (this work)} & Dialogue-note pairs: notes and dialogues role-played by LLMs & Dial-2-Note, Note-2-Dial & 1070/47 & 621/23 & 10,035 & Yes \\
    \bottomrule
  \end{tabular}
  % } % end of \small group if used
  \caption{Comparison of Dial-2-Note generation datasets. 'ASR': Automatic Speech Recognition.}
  \label{tab:dataset_comparison}
\end{table*}

\section{Methodology}
\label{sec:methodology}
%We introduce the data sources used, analyze the disease distribution, and discuss important quality factors identified through a comprehensive survey, which informs our dataset construction. The subsequent parts of this section describe the pipelines developed for generating medical notes and dialogues.
This section describes the methods taken to develop MedSynth, including an analysis of real-world disease distributions, important quality factors, and the novel data generation pipeline.

\subsection{Data Resource}
\label{subsec:data}
Four main data sources inform this research and are introduced in this section. 

%\subsubsection{IQVIA PharMetrics Plus} 
\textbf{IQVIA PharMetrics Plus} is a US medical insurance claims database, which has been the data source for several research findings such as cost savings associated with drug interactions \cite{yaghmaei2023causal}. We use this database to estimate the frequency of ICD-10 codes. We generate data for the top 2000 most frequent medical conditions. We had access to a 25\% random sample of the database from 2006 to 2021, which amounts to over 3 billion insurance claims. The database contains claims with either ICD-9 or ICD-10 code labels. ICD-10 codes were applied in practice in 2015 \cite{hirsch2016icd}. Mapping ICD-9 codes to ICD-10 is not straightforward due to significant structural and conceptual differences between the two systems. ICD-10 is not simply a superset of ICD-9 --- it introduces greater granularity \cite{cartwright2013icd} as well. %Therefore, mapping from ICD-10 to ICD-9 and vice versa is not one-to-one. 
Therefore, we filter the data to only include claims with ICD-10 codes. Moreover, according to Benes \citep{benes2023risk}, some codes are duplicated across the ICD-9 and ICD-10 systems. To avoid confusion, we remove the duplicates and end up with 800 million claim records and 54,985 unique ICD-10 codes.

%\subsubsection{Aci-Bench}
The \textbf{Aci-Bench} dataset contains 207 dialogue-note pairs and was generated by medical experts and lay persons role-playing \cite{yim2023aci}. As Table \ref{tab:dataset_comparison} presents, to the best of our knowledge, Aci-Bench is the largest publicly available dataset in terms of the length of the dialogues and notes that went through extensive human evaluation and is considered the latest benchmark for the medical Dial-2-Note and Note-2-Dial tasks and is publicly available under  Creative Commons Attribution 4.0 International License.

%\subsubsection{NoteChat}
\textbf{NoteChat} was introduced by Wang et al. \citep{wang2023notechat} and contains 207k dialogue-note pairs. 
The dialogues are synthetically generated by multiple cooperating role-playing LLMs. The notes come from PMC-Patient \cite{zhao2022pmc} available under CC BY-NC-SA license, which reported a wide coverage of MeSH\footnote{MeSH (Medical Subject Headings), used for indexing articles in PubMed.} disease terms. 
% (81.7\% of the total MeSH Disease Terms). 
% They reported MeSH since ICD codes were not relevant. 
NoteChat is one of the largest publicly available datasets in this area, in terms of the number of data points. 
However, its notes are not provided in standard medical formats but are instead patient summaries \cite{zhao2022pmc}, which limits NoteChat's applicability for Dial-2-Note and Note-2-Dial tasks, as observed in our experiments. 

%\subsubsection{PriMock57}
\textbf{PriMock57} contains 57 dialogue-note pairs generated through a simulation framework developed to mimic a telemedicine environment \cite{korfiatis2022primock57}. Trained physicians and laypersons were included in the study to simulate doctor-patient encounters and, after each interaction, physicians wrote notes in the SOAP format. 

\subsection{Diseases Distribution Analysis}
\label{subsec:disease_dist}

To ensure that our synthetic data reflects real-world disease prevalence, we analyze and identify the distributions of the most common diseases. To do so, we use the IQVIA PharMetrics Plus database.

Table \ref{tab:disease_dist} shows that disease distribution is highly skewed. We choose the most frequent 2000 ICD-10 codes and generate 5 dialogue-note pairs for each. We use this uniform sampling, despite the data skew, to prevent MedSynth from being dominated by common diseases, as our goal is to generate a synthetic dataset that can be used to train models that are robust over a diverse set of diseases. 
To the best of our knowledge, our dataset is the largest dataset of medical dialogue-note pairs available publicly in terms of the number and diversity of diseases that adhere to standard medical structure. Table \ref{tab:top_diseases} in Appendix \ref{sec:appendix_7} shows the 20 most prevalent ICD-10 codes in our analysis.

\begin{table}[!ht]
  \centering
  \begin{tabular}{l r}
    \toprule
    \textbf{\# of Claims} & \textbf{\# of ICD-10 Codes} \\
    \midrule
    Above 10M  & 4 \\
    1M - 10M   & 112 \\
    100k - 1M  & 1032 \\
    10k - 100k & 3922 \\
    1k - 10k   & 8942 \\
    Below 1k   & 40,973 \\
    \bottomrule
  \end{tabular}
  \caption{Distribution of ICD-10 codes from IQVIA Pharmetrics Plus database.}
  \label{tab:disease_dist}
\end{table}

\subsection{Survey on Important Quality Factors}
As mentioned by \cite{{mathioudakis2016keep}}, medical notes should adhere to standards to ensure continuity of care and facilitate communication between healthcare professionals. Good documentation should allow patients to access and understand the treatment procedure. Moreover, in case of an audit of healthcare services, medical notes can be a valuable resource. To ensure the synthetic medical notes capture useful information, particularly because real medical notes were inaccessible, we surveyed two medical professionals to determine the essential variables for note quality. One participant is a general physician and the other is a bio-statistician with over 13 years of experience with EHR systems and medical data. We listed 26 variables and asked the raters to score each variable between 0 and 10 in terms of how important each variable is in determining the quality of medical notes. In the survey, 0 represents the lowest importance, and 10 represents the highest importance. We then take the 10 variables with the highest average importance scores. There was a tie in the $10^{th}$ position, so we considered eleven variables initially. Appendix \ref{sec:appendix_5} discusses the results of the survey. Also, after obtaining the importance scores and discussing them with the medical experts, we added two more variables: Physical Exams and Investigations/Tests. 

We use the results from this survey and the disease distribution analysis to guide our data generation pipelines.

\subsection{Data Generation Pipeline}
\label{subsec:note_gin_pipeline}

We first generate notes and then use these notes to generate the corresponding dialogues. This is due to the fact that it is relatively easier to define "quality" in medical notes compared to dialogues, and then we can generate dialogues based on medical notes. Multiple LLM agents collaborate to generate these notes. We use GPT-4o in both note generation and dialogue generation pipelines. We use two prompting strategies: Chain-of-Thought (CoT) \cite{{wei2022chain}}  improves LLM performance in complex reasoning tasks through intermediate reasoning steps, and role-playing that acts as an effective trigger for the CoT process \cite{kong2023better}. All prompts are in Appendix \ref{sec:appendix_1}.

We also use In-Context Learning (ICL) as a training-free framework that does not require updating model parameters \cite{brown2020language}. ICL is directly applied to a pre-trained LLM and improves model performance by demonstrating good answers to queries without updating model parameters. In this paper, we use three forms of ICL:

\textbf{a) Few-Shot ICL:} At inference time, \( n \) demonstrations of the desired task are given to the model as conditioning in natural language.

\textbf{b) One-Shot ICL:} Few-Shot ICL with  $n=1$.

\textbf{c) Zero-Shot ICL:} Few-Shot ICL but a natural language description of the task is provided to the model instead of examples of the required task. We use this technique when there is no example available to provide to the model.

Previous research showed that model performance increases by increasing the number of examples in ICL \cite{agarwal2024many,brown2020language}. Therefore, we opt to use the maximum number of examples possible, while also considering computation costs. In this research, examples include dialogue-note pairs that translate to almost 1800 tokens per example. Therefore, we use three examples in the Few-Shot ICL setting 

\subsubsection{Note Generation Pipeline}
Figure \ref{fig:note_gen_pipeline} shows that our note-generation pipeline that consists of four agents: 

\begin{figure}[t]
  \includegraphics[width=\columnwidth]{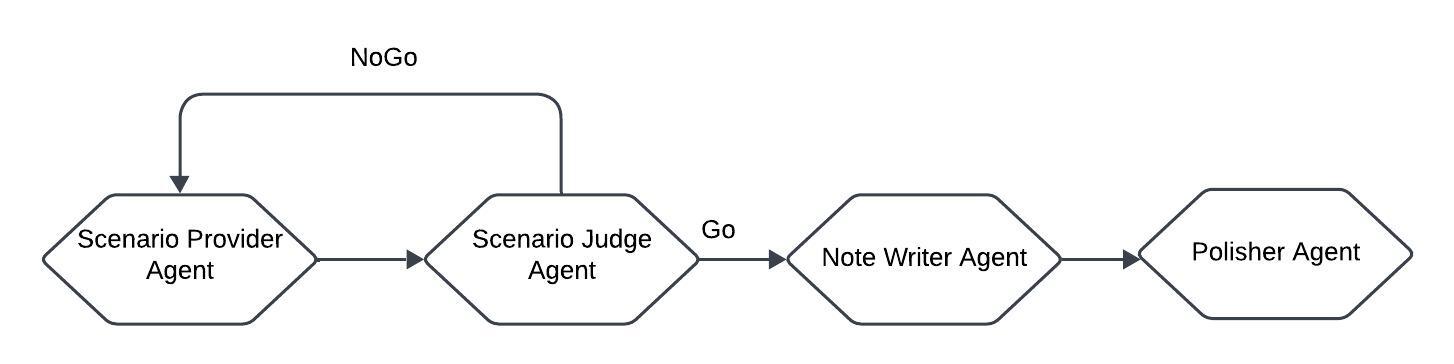}
  \caption{Note Generation Pipeline: The pipeline consists of four agents collaborating to generate the medical notes}
  \label{fig:note_gen_pipeline}
\end{figure}

\par
1) The \textbf{Scenario Provider Agent}
receives the disease description (ICD-10 description), selects an appropriate role based on the disease (e.g., if the disease involves a broken bone, the Scenario Provider chooses Orthopedist), and generates a scenario based on the thirteen specified variables that fit the disease description and the chosen role. 
In each generation, we sample a note from the Aci-Bench training set and provide it to the model.

2) The \textbf{Scenario Judge Agent} is tasked with evaluating the quality and validity of scenarios generated by the Scenario Provider Agent. Its main responsibility is to determine whether the provided scenarios meet specific conditions to be approved for generating synthetic medical notes. The evaluation process involves three checks:

\begin{enumerate}
\setlength{\itemsep}{0pt}
    \item Compare variables: The agent checks whether at least four out of the thirteen variables in the current scenario differ from previously approved scenarios for the same ICD-10 code. The threshold of four was determined empirically as providing the balance between computational complexity and variety of the notes. This ensures diversity and uniqueness in the generated scenarios.
    \item Medical Accuracy: The agent examines the scenario to confirm that it is medically accurate. This includes verifying that the symptoms, tests, diagnosis, and treatment are appropriate and logically consistent with the disease described by the ICD-10 code description.
    \item Plausibility: The agent evaluates the overall plausibility of the scenario to ensure it is realistic and could feasibly occur in a medical setting.
\end{enumerate}

If a scenario is rejected, detailed feedback is provided by the Scenario Judge to the Scenario Provider. The feedback contains textual suggestions from the Scenario Judge about what parts of the scenario could be improved to pass the checks. Appendix \ref{sec:appendix_6} includes an example of such feedback.
This loop continues until a pre-selected number of scenarios are approved. 

3) The \textbf{Note Writer Agent} receives a detailed scenario and the physician's role (e.g., a general physician, or a specialist in various fields). Table \ref{tab:role_frequency} in Appendix \ref{sec:appendix_3} shows the frequency statistics of the roles selected by the agent. Based on this scenario and the role, the agent generates a medical note adhering to the SOAP format. Details of the SOAP format are provided in Appendix \ref{sec:appendix_0}. 
The Note Writer Agent uses One-Shot ICL; in each generation, we sample a note from the Aci-Bench training set and provide it to the model to help generate notes in a similar format.

4) The \textbf{d) Note Polisher Agent} is designed to refine and enhance synthetic medical notes. The primary task is to ensure that the information in the notes is correctly categorized and placed in the appropriate sections.

\subsubsection{Dialogue Generation Pipeline}
\label{subsec:dial_gen_pipeline}

Our dialogue generation pipeline consists of two agents: the Dialogue Generator and the Dialogue Polisher.

\textbf{a) The Dialogue Generator Agent} generates a dialogue relevant to a given medical note. We use Few-Shot ICL with $n=3$ and feed three examples from the Aci-Bench training set to boost the quality of the generation.

\textbf{b) The Dialogue Polisher Agent} enhances and expands doctor-patient conversations by including social chatter to make them more realistic. More importantly, it double-checks to ensure all information from the medical note is accurately included in the dialogue. This is done 
to reduce hallucinations in the generated dialogue.

%\section{Experimental Details and Results}
\section{Experiments}
In this section, we analyze MedSynth's characteristics and assess the extrinsic quality of MedSynth by testing it on Dial-2-Note and Note-2-Dial tasks, demonstrating its effectiveness in training SotA models. All the experiments are run on a single A40 GPU and models are 4-bit quantized for computational efficiency.

\subsection{Data Characteristics}
Table \ref{tab:Medsynth_characteristic} shows the statistical characteristics of MedSynth. It covers 2001 unique ICD-10 codes and over 10,000 dialogue-note pairs. 

\begin{table}[!ht]
  \centering
  \begin{tabular}{l r}
    \toprule
    \textbf{Metric} & \textbf{MedSynth} \\
    \midrule
    \# of data points & 10,035 \\
    \# of unique ICD-10 codes & 2001 \\
    \midrule
    \textbf{Dialogue} & \\
    \midrule
    Avg length  (tok) & 932 \\
    Avg length (sentences) & 55 \\
    \midrule
    \textbf{Note} & \\
    \midrule
    Avg length (tok) & 621 \\
    Avg length (sentences) & 23 \\
    \bottomrule
  \end{tabular}
  \caption{Statistical characteristics of MedSynth. Note `tok' does not include speaker IDs.}
  \label{tab:Medsynth_characteristic}
\end{table}

\subsection{MedSynth for the Dial-2-Note and Note-2-Dial Generation}
Human evaluation of generated data is costly and time-consuming \cite{{liu2024towards}}. Therefore, we %take an automatic approach to assessing 
automatically assess the quality and usefulness of the data for Dial-2-Note and Note-2-Dial generation. Aci-Bench represents the latest benchmark for these tasks. The central hypothesis is that, if MedSynth's quality is high and its dialogues and notes closely resemble the distribution of Aci-Bench, then models fine-tuned on MedSynth should achieve strong performance on the Aci-Bench test set. We run a series of experiments to extrinsically evaluate the quality of MedSynth.% across two tasks: Dial-2-Note and Note-2-Dial. 

\subsubsection{Evaluation Metrics}

Recently research suggests that traditional metrics such as BLEU \cite{papineni2002bleu} and ROUGE \cite{lin2004rouge} do not correlate well with human judgment in evaluating machine-generated text and, therefore, do not capture the depth and granularity of human evaluation \cite{zhang2019bertscore,krishna2021hurdles}. Using LLMs as judges has become a prevalent alternative due to their potential parity with human evaluation \cite{zheng2023lmsys}. Therefore, we use this setting as our primary evaluation strategy. We also report traditional metrics, i.e., ROUGE \cite{lin2004rouge}, BLEU \cite{papineni2002bleu}, and METEOR \cite{banerjee2005meteor} in Appendix \ref{sec:appendix_00}.

Verga et al. \citep{verga2024replacing} reported that "juries" composed of diverse LLMs can outperform a single judge and
decrease intra-model bias. % due to its composition of disjoint model families.
Therefore, due to the sensitivity of medical applications such as our targeted tasks, we %follow the same setting and 
use majority voting with three LLM judges from different model families: Prometheus \cite{kim2023prometheus}, GPT-4o, and Qwen2.5-32B-Instruct \cite{yang2024qwen2}. We also perform human evaluation with three medical professionals to assess the validity of the LLM jury setting. 

In our case, we provide the reference note, the notes generated by two models, and a rubric to each Judge LLM and ask it to decide which model performed best according to the rubric and reference note. Then, we use majority voting to decide which note is chosen. For the Dial-2-Note task, the rubric includes seven aspects: Hallucination, Critical Omissions, Professional Tone, Logical Structure of the Note and Sentences, Adherence to the SOAP Format, and Section Relevance. For the Note-2-Dial task, the rubric includes: Completeness, Accuracy, Naturalness and Flow, Use of Medical Terminology, and Evidence-Based Support. For both tasks, these criteria are selected based on previous research for the same tasks where researchers tried to gather human-expert feedback \cite{abacha2023empirical,wang2023notechat}. All the prompts and the detailed rubric can be found in Appendix \ref{sec:appendix_1}.

\subsubsection{Results: Dial-2-Note Task}
We sample data from NoteChat to match the sample size of MedSynth, combine these with the Aci-Bench training set, and fine-tune the Llama-3-8B-Instruct model. Finally, we test the tuned models on the Aci-Bench test set. To further evaluate the value of MedSynth, we run an additional experiment where the Aci-Bench training set is excluded from training to compare the standalone quality of MedSynth against NoteChat. We repeat the experiments to contrast MedSynth with PriMock57 instead of with NoteChat. Moreover, we contrast the model tuned on MedSynth and Aci-Bench's training set with the model solely trained on the Aci-Bench training set as a baseline. Table \ref{tab:d2n_llm_eval_results} shows the results of these experiments.

In all evaluations, when MedSynth is included in the training set, the corresponding model is preferred by the jury to any other model. We also observe that excluding the Aci-Bench training set from the training data widens the performance gap in favour of MedSynth. Results of the traditional metrics reported in Appendix \ref{sec:appendix_00} are also consistent with these observations. 

An important observation is that when the models were fine-tuned exclusively on MedSynth compared with exclusively on NoteChat, a significant disparity in performance occurs in their adherence to the SOAP format. Specifically, the model fine-tuned only on NoteChat failed to maintain the structured SOAP format, which is a critical component of standard medical documentation. This confirms our expectation because NoteChat only includes patient summaries instead of structured medical notes. An example of such a failure is provided in Appendix \ref{sec:appendix_4}.

\begin{table}
  \centering
  \small
  \begin{tabular}{p{0.11\textwidth} | r | r}
    \toprule
    \textbf{Comparison} & \textbf{Num. samples} & \textbf{Jury Pref. Rate}  \\
    \midrule
    MS+AT vs. NC+AT & 10,102 &   60.0\% \\
    \midrule
    MS Only vs. NC Only & 10,035 & 95.0\% \\
    \midrule
     MS+AT vs. AT Only & 10,102 &  52.5\% \\
    \midrule
    MS+AT vs. PM+AT & 124 & 57.5\%  \\
    \midrule
    MS Only vs. PM Only & 57 & 90.0\% \\
    \bottomrule
  \end{tabular}
  \caption{LLM-based evaluation results on Aci-Bench for Dial-2-Note task. 'MS': MedSynth; 'AT': Aci\_Train; 'PM': PriMock; 'NC': NoteChat; 'Jury Pref. Rate': Jury Preference Rate \textbf{in favour of MedSynth}. Note that the column \textbf{Num. samples} shows the number of samples in the training set.}
  \label{tab:d2n_llm_eval_results}
\end{table}

\subsubsection{Results: Note-2-Dial Task}
In the Note-2-Dial task, we use the same setting as the Dial-2-Note task and contrast MedSynth against NoteChat and PriMock57. We fine-tuned the Llama-3-8B-Instruct model under various conditions: one where the Aci-Bench training set was included alongside MedSynth and NoteChat, and another where it was excluded.  The models were then evaluated on the Aci-Bench test set. We repeated the experiments with PriMock57 instead of NoteChat to further assess MedSynth's usefulness. Table \ref{tab:note2dial_llm_eval} shows the results.

Similar to the Dial-2-Note task, the results show that including MedSynth in the training set leads to improved performance across all our experiments. Moreover, as observed in the Dial-2-Note task, excluding the Aci-Bench training set from training leads to bigger gaps in performance in favor of MedSynth. The results of the traditional metrics presented in Table \ref{tab:stats_eval_results_note2dial} in Appendix \ref{sec:appendix_00} are consistent with these observations. 

Although MedSynth uses a simpler dialogue-generation pipeline compared to NoteChat, it outperforms NoteChat in generating dialogues from notes. This could be associated with the effectiveness of Few-Shot ICL in generating high-quality dialogues compared to the multi-agent setting without Few-Shot ICL used in NoteChat.

\begin{table}[!ht]
  \centering
  \small
  \begin{tabular}{p{0.11\textwidth} | r | r}
    \toprule
    \textbf{Comparison} & \textbf{Num. samples} & \textbf{Jury Pref. Rate} \\
    \midrule
    \shortstack[l]{MS+AT vs.\\ NC+AT}   &   10,102 & 55.0\% \\
    \midrule
    \shortstack[l]{MS Only vs.\\ NC Only}   & 10,035 & 87.5\% \\
    \midrule
    \shortstack[l]{MS+AT vs.\\ AT Only} & 10,102 &  80.0\% \\
    \midrule
    
     \shortstack[l]{MS+AT vs.\\ PM+AT}  & 124 & 57.5\% \\
    \midrule
    \shortstack[l]{MS Only vs.\\ PM Only} & 57 & 97.5\% \\
    \bottomrule
  \end{tabular}
  \caption{LLM-based evaluation results for Note-2-Dial task. 'MS': MedSynth; 'AT': Aci\_Train; 'NC': NoteChat; 'PM': PriMock; 'Jury. Pref. Rate': Jury Preference Rate \textbf{in favor of MedSynth.} Note that the column \textbf{Num. samples} shows the number of samples in the training set.}
  \label{tab:note2dial_llm_eval}
\end{table}

\subsection{Human Evaluation Results}
To validate LLM-based evaluation, we conducted a human study on one of the most competitive comparison (MS+AT vs NC+AT) in the Dial2Note task. Three medical professionals evaluated 40 samples with good inter-annotator agreement (ICC=0.75). 

When determining the preferred candidate by majority vote per sample, both approaches performed equally (50\% each, 20-20). However, when considering individual votes, annotators showed a slight preference for MedSynth (54.2\%, 65/120 votes). This aligns with the LLM jury's 60\% preference, suggesting both humans and LLMs prefer MedSynth-trained models. Table \ref{tab:human_eval} presents the results of the human evaluation.
\begin{table}[!ht]
  \centering
  \small
\begin{tabular}{p{0.14\textwidth} | p{0.08\textwidth} | >{\raggedleft\arraybackslash}p{0.08\textwidth} | >{\raggedleft\arraybackslash}p{0.08\textwidth}}
    \toprule
    \textbf{Evaluation} & \textbf{Metric} & \textbf{MS+AT} & \textbf{NC+AT} \\
    \midrule
    LLM Jury & Samples preferred & 60.0\% & 40.0\% \\
    \midrule
    \shortstack[l]{Human Experts\\(majority vote)} & Samples won & 50.0\% & 50.0\% \\
    \midrule
    \shortstack[l]{Human Experts\\(individual votes)} & Vote share & 54.2\% & 45.8\% \\
    \bottomrule
  \end{tabular}
  \caption{Comparison of LLM and human evaluation for MS+AT vs.\ NC+AT. Three medical professionals evaluated 40 samples with good inter-annotator agreement (ICC(2,k)=0.75). While the majority vote shows a tie, individual vote analysis reveals a slight preference for MedSynth, consistent with LLM evaluation.}
  \label{tab:human_eval}
\end{table}

\section{Ablation Studies}
We conduct a series of ablation studies to assess components of our proposed pipeline. First, we ablated the Scenario Judge Agent by once removing it and once swapping the backbone LLM from GPT-4o to Llama-3.3-70B-Instruct. Second, we swapped all backbone LLMs in the whole data generation pipeline with either of two other models: once with Llama-3.3-70B-Instruct and once with Qwen2.5-32B-Instruct \cite{yang2024qwen2}. We generated 246 samples in each of these configurations with the same ICD-10 codes. We repeat the experiments for both Dial-2-Note and Note-2-Dial tasks. All the experiments are run on a single A40 GPU and models are 4-bit quantized for computational efficiency.
Additional experiments are provided in Appendix \ref{sec:appendix_more_experiments}
\subsection{Ablation: Dial-2-Note Task}.

The results of the ablations for the Dial-2-Note task are presented in Table \ref{tab:ablation_D2N}, and traditional metrics are reported in Table \ref{tab:stats_eval_results_ablation_dial_2_note} in the Appendix \ref{sec:appendix_00}. By comparing the performance of the model tuned on the data generated by removing the Scenario Judge Agent from the pipeline (NuJudge+AT) with the one with the agent present (AllGPT+AT), the first observation is that the Scenario Judge Agent is effective in the whole pipeline for the Dial-2-Note task. The win rate of AllGPT+AT over NuJudge+AT was 57.5\%. However, the agent seems to be sensitive to the backbone LLM since swapping it to Llama-3.3-70B-Instruct (Lamma-JudgeGPT+AT) led to weaker performance. The win rate of NuJudge+AT over Lamma-JudgeGPT+AT was 52.5\%. Traditional metrics point in the same direction by showing lower scores for Lamma-JudgeGPT+AT. %However, the scores were mostly the same for AllGPT+AT and NuJudge+AT.

The second observation is that not using GPT-4o as the backbone model decreases  performance, especially compared to NoteChat. This might be due to the fact that NoteChat was created using a multi-agent LLM collaboration, with the GPT family of models as the backbone. Further evaluation with both pipelines, using the same backbone family of models, might shed more light on the contribution of that architecture.

\begin{table}[!ht]
\small
  \centering
  \begin{tabular}{
    p{0.15\textwidth}
    | >{\centering\arraybackslash}p{0.045\textwidth}
    | >{\raggedright\arraybackslash}p{0.16\textwidth}
  }
    \toprule
    \textbf{Comparison} & \textbf{Num. samples} & \textbf{Jury Pref. Rate}  \\
    \midrule
     \shortstack[l]{NoJudge+AT\\ vs.\\ AllGPT+AT} & 313 & \shortstack[r]{57.5\% \\ \small\textbf{(AllGPT+AT)}} \\
    \midrule
     \shortstack[l]{NoJudge+AT\\ vs.\\ LammaJudgeGPT\\+AT} & 313 &\shortstack[r]{52.5\%\\ \small\textbf{(NoJudge+AT)}} \\
    \midrule
    \midrule
    
    \shortstack[l]{AllLamma+AT\\ vs.\\ PM+AT} &  124 & \shortstack[r]{52.5\% \\ \small\textbf{(AllLamma+AT)}} \\
    \midrule
     \shortstack[l]{AllQwen+AT\\ vs.\\ PM+AT} &  124 & \shortstack[r]{52.5\%\\ \small \textbf{(PM+AT)}}   \\
    \midrule
    \shortstack[l]{AllLamma+AT\\ vs.\\ NC+AT}& 313 & \shortstack[r]{55.0\%\\ \small \textbf{(NC+AT)}} \\
    \midrule
    \shortstack[l]{AllQwen+AT\\ vs.\\ NC+AT}& 313 & \shortstack[r]{60.0\%\\ \small \textbf{(NC+AT)}}\\
    \bottomrule
  \end{tabular}
  \caption{LLM-based evaluation results on Aci-Bench for ablation studies on Dial-2-Note task. 'MS': MedSynth; 'AT': Aci\_Train; 'PM': PriMock; 'NC': NoteChat; 'Jury Pref. Rate': Jury Preference Rate (Winner). Note that the column \textbf{Num. samples} shows the number of samples in the training set.}
  \label{tab:ablation_D2N}
\end{table}

\subsection{Ablation: Note-2-Dial Task}
Table \ref{tab:ablation_N2D} shows that removing the Scenario Judge Agent and swapping the backbone model has different effects on the Note-2-Dial and Dial-2-Note tasks. This suggests that the pipeline might be sensitive to the selection of the backbone model. By swapping all backbone models, we observe consistent performance changes with the Dial-2-Note task, suggesting that the pipeline seems to best perform with GPT-4o as the backbone. Further evaluation with a single family of models across both the MedSynth and NoteChat pipelines is suggested for future work. 

\begin{table}[!ht]
\small
  \centering
  \begin{tabular}{
    p{0.15\textwidth}
    | >{\centering\arraybackslash}p{0.045\textwidth}
    | >{\raggedright\arraybackslash}p{0.24\textwidth}
  }
    \toprule
    \textbf{Comparison} & \textbf{Num. samples} & \textbf{Jury Pref. Rate}  \\
    \midrule
     \shortstack[l]{NoJudge+AT vs.\\ AllGPT+AT} & 313 & \shortstack[r]{55.0\%\\ \small\textbf{(NoJudge+AT)}}  \\
    \midrule
    \shortstack[l]{ NoJudge+AT vs.\\ LammaJudgeGPT \\ +AT} & 313 &
     \shortstack[r]{55.0\% \\\small\textbf{ (LammaJudgeGPT+AT)}} \\
    \midrule
    \midrule
    
     \shortstack[l]{AllLamma+AT vs.\\ PM+AT} &  124 & \shortstack[r]{60.0\%\\ \small \textbf{(PM+AT)}} \\
    \midrule
     \shortstack[l]{AllQwen+AT vs.\\ PM+AT} &  124 & \shortstack[r]{55.0\%\\ \small\textbf{(AllQwen+AT)}}   \\
    \midrule
     \shortstack[l]{AllLamma+AT vs.\\ NC+AT}& 313 & \shortstack[r]{67.5\%\\ \small \textbf{(NC+AT)}}  \\
    \midrule
     \shortstack[l]{AllQwen+AT vs.\\ NC+AT}& 313 & \shortstack[r]{57.5\% \\\small \textbf{(NC+AT)}}  \\
    \bottomrule
  \end{tabular}
  \caption{LLM-based evaluation results on Aci-Bench for ablation studies on Note-2-Dial task. 'MS': MedSynth; 'AT': Aci\_Train; 'PM': PriMock; 'NC': NoteChat; 'Jury Pref. Rate': Jury Preference Rate (Winner). Note that the column \textbf{Num. samples} shows the number of samples in the training set.}
  \label{tab:ablation_N2D}
\end{table}

% \subsection{Additional Experiments}

\section{Discussion}

We introduce MedSynth, a novel synthetic dataset designed to advance dialogue-to-note generation. MedSynth appears to be the first dataset of fully synthetic medical notes and dialogues, a crucial innovation in a field where data is scarce due to stringent privacy concerns.  By incorporating over 2000 ICD-10 codes and more than 10,000 dialogue-note pairs, MedSynth overcomes the absence of diverse and privacy-compliant datasets for medical SOAP note generation. Although the focus of MedSynth is on primary care, it also contains diseases related to other specialties and can be used in any context that follows the SOAP format.

We show that models trained on MedSynth exhibit improvements in generating accurate and contextually appropriate medical notes from dialogues, which is essential for reducing the documentation burden on healthcare providers and addressing physician burnout. 
In contrast to prior datasets, MedSynth provides SOAP-structured clinical notes aligned with a primary-care–like condition distribution, and uses insurance claims-informed sampling to broaden condition and comorbidity diversity at scale.
% Existing datasets, such as Aci-Bench and NoteChat, though valuable, are limited either in the number of data points, the diversity of medical conditions covered, or in the (lack of) structure of medical notes. Aci-Bench, for instance, is limited in size with only 207 dialogue-note pairs and NoteChat uses patient summaries rather than structured medical notes. 

We release our fine-tuned models that achieve SotA performance in the Dial-2-Note and Note-2-Dial tasks, which can be used as base models for future tool development.

\section{Limitations and Ethical Considerations}

First, while our pipeline is effective in generating synthetic data to improve Dial-2-Note and Note-2-Dial models, it does not guarantee the medical correctness of the generated text. Further research and expert evaluation are necessary to ensure medical accuracy. Therefore, MedSynth should not be considered a source of reliable medical information but rather a tool for model development \cite{walonoski2018synthea}.

Second, the synthetic nature of MedSynth, though beneficial for privacy, may not fully capture the complexities of real-world clinical dialogues between humans. Future work should evaluate the realism of these dialogues, to account for real-world sources of noise, and expand the dataset to include a broader range of diseases beyond the top 2000 ICD-10 codes, increasing its overall utility.

Third, although SOAP is among the most widely used structures for recording medical notes \cite{podder2022soap}, other structures exist. Future work should consider covering more structures of medical notes.

Finally, while our results improve model performance in controlled settings, it is crucial to practice safe, careful `MLOps' techniques when evaluating these models in clinical environments \cite{khattak2024mlhops}. Validating MedSynth’s impact on documentation efficiency and accuracy in real-world clinical workflows, with `real world' indicators of performance, will be an essential next step.

\bibliography{custom}

\appendix

\newpage
\appendix
\section{Statistical Performance Metrics}
\label{sec:appendix_00}

This section reports the traditional statistical metrics used for model evaluation. All the metrics in this section are used from \textbf{evaluate} library from \textbf{hugggingface}. 

\begin{table*}

  \centering
  \tiny 
  \sisetup{round-mode=places,round-precision=2}
  \begin{tabular}{>{\raggedright\arraybackslash}p{0.08\textwidth} |>{\raggedright\arraybackslash}p{0.17\textwidth} |>{\raggedright\arraybackslash}p{0.09\textwidth} |>{\raggedright\arraybackslash}p{0.09\textwidth} |>{\raggedright\arraybackslash}p{0.09\textwidth} |>{\raggedright\arraybackslash}p{0.09\textwidth} |>{\raggedright\arraybackslash}p{0.09\textwidth} |>{\raggedright\arraybackslash}p{0.09\textwidth}}
    \toprule
    \textbf{Samples} & \textbf{Dataset} & \textbf{BLEU} & \textbf{ROUGE-1} & \textbf{ROUGE-2} & \textbf{ROUGE-L} & \textbf{ROUGE-LSum} & \textbf{METEOR}  \\
    \midrule
    67 & AT & \num{0.1582616331978378} \tiny(\num{0.07684092223938284}) & \num{0.47659633411322194} \tiny(\num{0.14872065497969078}) & \num{0.2296182074079315} \tiny(\num{0.09099126169327523})  & \num{0.2865023879201488} \tiny(\num{0.1048843769963567})  & \num{0.44007303048437435}  \tiny(\num{0.14086921144020484})  &  \num{0.3633938382474824} \tiny(\num{0.07998620280575293})  \\
    \midrule
    10102 & MS+AT & \underline{\num{0.2806907251797213}} \tiny(\num{0.07625696836080391})  & \underline{\num{0.615070087928188}} \tiny(\num{0.06378846277577964})  & \underline{\num{0.3452563183456675}} \tiny(\num{0.07373369315009343})  & \underline{\num{0.4080639098035033}} \tiny(\num{0.07767291736014913})  & \underline{\num{0.5708353877009628}} \tiny(\num{0.060812918698586495})  & 
    \num{0.469914411796133} \tiny(\num{0.07945239500144312})  \\
    
    10102 & NC+AT & \num{0.273885610324215} \tiny(\num{0.08153099843095149}) & \num{0.6049955813513186} \tiny(\num{0.09724812835871485}) & \num{0.3398303636150718}  \tiny(\num{0.08860608940724078}) 
    & \num{0.3933856594474839} \tiny(\num{0.09891471045620538}) & 
    \num{0.557168786206413} \tiny(\num{0.09190118155033893}) & 
    \num{0.4666645453368477} \tiny(\num{0.08054436854787471}) \\
    \midrule
    10035 & MS & \underline{\num{0.11506082336162371}}\tiny(\num{0.036106717570498624})  & 
    \underline{\num{0.517016935060228}} \tiny(\num{0.04860355378691553}) & \underline{\num{0.22673439259784384}}\tiny(\num{0.04912258222903875})  & \underline{\num{0.29750128080827754}} \tiny(\num{0.05654220078386763})& \underline{\num{0.4837862906421071}} \tiny(\num{0.04941429252369985}) & \underline{\num{0.3812441247168886}} \tiny(\num{0.056153533113123384}) \\
    
    10035 & NC & \num{0.08726225220244274} \tiny(\num{0.05050698780596295}) & \num{0.3790693071751007}\tiny(\num{0.12295497027531521}) & 
    \num{0.1546660211833093} \tiny(\num{0.07314038684870165}) & \num{0.2207404778687562} \tiny(\num{0.08332822396923352}) & \num{0.33396388335475713} \tiny(\num{0.10802517050152644}) &  \num{0.33513209499554775} \tiny(\num{0.08070987274368273}) \\

        \midrule
    124 & MS+AT & \underline{\num{0.19500809579280073}} \tiny(\num{0.08698583455837383})  & 
    \underline{\num{0.540797784473949}} \tiny(\num{0.09139040422143144})  & \underline{\num{0.26918458528873085}} \tiny(\num{0.08365848063047299})  & \underline{\num{0.32754964777832896}} \tiny(\num{0.08292509010556273})  & \underline{\num{0.5003801498609004}} \tiny(\num{0.08812287946147714})  & 
\underline{\num{0.39663855225630085}} \tiny(\num{0.09728374773263909})  \\

    124 & PM+AT & \num{0.17277547229838602} \tiny(\num{0.0748621820678834})  & \num{0.516554036703367} \tiny(\num{0.09579187414984305})  & \num{0.25190813815149615} \tiny(\num{0.074469103440148})  & \num{0.31546300425635315} \tiny(\num{0.08048555997770115})  & \num{0.4786755789779914} \tiny(\num{0.08962869617379003})  & 
    \num{0.38781476268961096} \tiny(\num{0.07929953194015524})  \\
    
               \midrule
               
    57 & MS & \underline{\num{0.09650340144664729}} \tiny(\num{0.04823427150954164})  & \underline{\num{0.4729278360178979}} \tiny(\num{0.08858171441462344})  & \underline{\num{0.19533107262792693}} \tiny(\num{0.06825451802407444})  & \underline{\num{0.2593202518310401}} \tiny(\num{0.07125560534442217})  & \underline{\num{0.4404600263300381}} \tiny(\num{0.08540847179578087})  & 
    \underline{\num{0.3392910758887551}} \tiny(\num{0.056454822301508896})  \\

    57 & PM & \num{0.022980908624033302}
    \tiny(\num{0.03657556635356276})  & \num{0.17033861535574552}
    \tiny(\num{0.16138397692072426})  & \num{0.06630507623119952}\tiny(\num{0.07663252417230926})  & \num{0.11157253095812347}
    \tiny(\num{0.1047974460395766})  & \num{0.15289418647437303}
    \tiny(\num{0.14375905348053297})  & 
    \num{0.1169240531185088} \tiny(\num{0.0949469725055781})  \\
    \bottomrule
  \end{tabular}
  \caption{Comparison of model performance by dataset and sample size for Dial-2-Note task. 'MS+AT': MedSynth+Aci\_Train; 'NC+AT': NoteChat+Aci\_Train; 'PM+AT': PriMock57+Aci\_Train. Note that the column \textbf{Samples} shows the number of samples in the training set.}
  \label{tab:stats_eval_results_dial2noteee}
\end{table*}

\begin{table*}
  \centering
  \tiny 
  \sisetup{round-mode=places,round-precision=2}
  \begin{tabular}{>{\raggedright\arraybackslash}p{0.08\textwidth} |>{\raggedright\arraybackslash}p{0.17\textwidth} |>{\raggedright\arraybackslash}p{0.09\textwidth} |>{\raggedright\arraybackslash}p{0.09\textwidth} |>{\raggedright\arraybackslash}p{0.09\textwidth} |>{\raggedright\arraybackslash}p{0.09\textwidth} |>{\raggedright\arraybackslash}p{0.09\textwidth} |>{\raggedright\arraybackslash}p{0.09\textwidth}}
    \toprule
    \textbf{Samples} & \textbf{Dataset} & \textbf{BLEU} & \textbf{ROUGE-1} & \textbf{ROUGE-2} & \textbf{ROUGE-L} & \textbf{ROUGE-LSum} & \textbf{METEOR}  \\
    \midrule
    67 & AT & \num{0.08400418382738528} \tiny(\num{0.05299095759655446}) & \num{0.36584641507448695} \tiny(\num{0.1399511365835181}) & \num{0.13785486428199314} \tiny(\num{0.07302177232933006}) & \num{0.18200589757118485} \tiny(\num{0.05460118305520479}) & \num{0.3378944207007311} \tiny(\num{0.13294455443609116}) &  \num{0.2627117531839702} \tiny(\num{0.07533172142257072})\\
    \midrule
    
    10102 & MS+AT & \underline{\num{0.1384617116277671}} \tiny(\num{0.07425100179668169})& \underline{\num{0.5413457815811412}} \tiny(\num{0.09985887808144149})& \num{0.24184370599334057} \tiny(\num{0.0692758489350449})& \underline{\num{0.2550102068513201}} \tiny(\num{0.06309038973086255})& \underline{\num{0.511747881584028}} \tiny(\num{0.09726686562800052})&  \underline{\num{0.3280906419668105}} \tiny(\num{0.09186907380010374}) \\
    
    10102 & NC+AT & \num{0.1150811384601462} \tiny(\num{0.06495503704270111})& \num{0.5124788561467024} \tiny(\num{0.10652440689060642})& \num{0.2350811807823666} \tiny(\num{0.06163714680716085})& \num{0.2482343459886247} \tiny(\num{0.06539258169448564})& \num{0.48694230599270955} \tiny(\num{0.10204471316282783})& \num{0.28992200429867043} \tiny(\num{0.07385623010426685}) \\
    \midrule
    10035 & MS & \underline{\num{0.08775218153419342}} \tiny(\num{0.043255208593592644})& \underline{\num{0.5111494586339512}} \tiny(\num{0.06634467308811663})& \underline{\num{0.19298178898340126}} \tiny(\num{0.029142120660376723})& \underline{\num{0.22874657529849554}} \tiny(\num{0.0432727277650609})& \underline{\num{0.4874052445900042}} \tiny(\num{0.06355148998578752})& \underline{\num{0.2912102618333778}} \tiny(\num{0.06359813990861055}) \\
    
    10035 & NC & \num{0.0328378006721944} \tiny(\num{0.02260731144572997})& \num{0.4286867834227538} \tiny(\num{0.07808001779945829})& \num{0.17322750787474614} \tiny(\num{0.04220118701826139})& \num{0.21935487712132123} \tiny(\num{0.054534070556955354})& \num{0.4089608731217025} \tiny(\num{0.07709703846437148})&  \num{0.18806213898229399} \tiny(\num{0.049475040725241275})\\
            \midrule
    124 & MS+AT & \underline{\num{0.14598490943985623}} \tiny(\num{0.06583307481776454})  & 
    \underline{\num{0.530855993733644}} \tiny(\num{0.08809585661247736})  & \underline{\num{0.22722274325038563}} \tiny(\num{0.04948837609994322})  & \underline{\num{0.24393053242099388}} \tiny(\num{0.04957591350398667})  & \underline{\num{0.503400490160826}} \tiny(\num{0.08563167909578577})  & 
\underline{\num{0.3205401802663295}} \tiny(\num{0.06749655735088222})  \\

    124 & PM+AT & \num{0.12043019013720353} \tiny(\num{0.06551701190250522})  & \num{0.4721851122682569} \tiny(\num{0.13144797653210707})  & \num{0.18338432318452766} \tiny(\num{0.07261225982715411})  & \num{0.2189640733206067} \tiny(\num{0.05503463312904503})  & \num{0.4412239204173499} \tiny(\num{0.129829701504194})  & 
    \num{0.30928707990431487} \tiny(\num{0.0846940653279491})  \\
    
               \midrule
               
    57 & MS & \underline{\num{0.08859216627894453}} \tiny(\num{0.04862021593785226})  & \underline{\num{0.512330622089299}} \tiny(\num{0.0781180495869028})  & \underline{\num{0.19410317442695968}} \tiny(\num{0.031119113966450236})  & \underline{\num{0.22562330910609432}} \tiny(\num{0.04200961312685592})  & \underline{\num{0.4892257410205566}} \tiny(\num{0.07477107909523895})  & 
    \underline{\num{0.30010351738731744}} \tiny(\num{0.08098592480041265})  \\

    57 & PM &\num{0.010330219071088045} \tiny(\num{0.01199076338096373})  & \num{0.16555593624911655}
    \tiny(\num{0.13166621440354595})  & \num{0.039281058187103736} \tiny(\num{0.03657558183813358})  & \num{0.09558768763174345} \tiny(\num{0.06898262506901875})  & \num{0.15145113581928094} \tiny(\num{0.12027780067799838})  & 
    \num{0.1327759451253939} \tiny(\num{0.10176682071515251}) \\
    \bottomrule
  \end{tabular}
  \caption{Comparison of model performance by dataset and sample size for Note-2-Dial task. 'MS+AT': MedSynth+Aci\_Train; 'NC+AT': NoteChat+Aci\_Train; 'PM+AT': PriMock57+Aci\_Train. Note that the column \textbf{Samples} shows the number of samples in the training set.}
  \label{tab:stats_eval_results_note2dial}
\end{table*}

\begin{table*}
  \centering
  \tiny 
  \sisetup{round-mode=places,round-precision=2}
  \begin{tabular}{>{\raggedright\arraybackslash}p{0.08\textwidth} |> {\raggedright\arraybackslash}p{0.17\textwidth} |>{\raggedright\arraybackslash}p{0.09\textwidth} |>{\raggedright\arraybackslash}p{0.09\textwidth} |>{\raggedright\arraybackslash}p{0.09\textwidth} |>{\raggedright\arraybackslash}p{0.09\textwidth} |>{\raggedright\arraybackslash}p{0.09\textwidth} |>{\raggedright\arraybackslash}p{0.09\textwidth}}
    \toprule
    \textbf{Samples} & \textbf{Dataset} & \textbf{BLEU} & \textbf{ROUGE-1} & \textbf{ROUGE-2} & \textbf{ROUGE-L} & \textbf{ROUGE-LSum} & \textbf{METEOR}  \\
    
    \midrule
      313 & NoJudge+AT & \underline{\num{0.25560564813448583}} \tiny(\num{0.0870704010728623})& \underline{\num{0.5905881226098153}} \tiny(\num{0.08239045368633463})& \underline{\num{0.32464656453842816}} \tiny(\num{0.08580886777359027})& \underline{\num{0.3816310147626677}} \tiny(\num{0.08724423677062126})& \underline{\num{0.5455847455985416}} \tiny(\num{0.08724187494617719})&  \underline{\num{0.4502819152574447}} \tiny(\num{0.08443499348549907})\\
      
      313 & AllGPT+AT & \underline{\num{0.25554811516135756}} \tiny(\num{0.08797821161670691})& \num{0.5832117566999099}
      \tiny(\num{0.1108560256928654})& \underline{\num{0.32033931054829373}} \tiny(\num{0.0937654108646255})& \num{0.3737301376133201} \tiny(\num{0.09900982825842933})& \num{0.5385024967028744} \tiny(\num{0.10504884071345148})&  \underline{\num{0.44615619420110714}} \tiny(\num{0.08889945311831074})\\
    
     313 & LammaJudgeGPT+AT & \num{0.22952874676727886} \tiny(\num{0.08828300644183809})& \num{0.5663577422811961} \tiny(\num{0.09274213503192197})& \num{0.3056037524604419} \tiny(\num{0.08524977488426132})& \num{0.3624297551581753} \tiny(\num{0.09068943400431619})& \num{0.5254070078812548} \tiny(\num{0.08875126807477045})& \num{0.42406559544000844} \tiny(\num{0.09387158696784241})\\
    \midrule
    
    124 & AllLamma+AT & \num{0.17582675526016525} \tiny(\num{0.07993496988350939})& \num{0.5043688705994697} \tiny(\num{0.12313895314151166})& \num{0.24556084074289966} \tiny(\num{0.08608606924403087})& \num{0.29598845451461814} \tiny(\num{0.09108646010746357})& \num{0.4601804046049306} \tiny(\num{0.112875324327709})& \num{0.3890412542715964} \tiny(\num{0.08130143954105817})\\
    
     124 & AllQwen+AT & \underline{\num{0.19519295547389678}} \tiny(\num{0.06890639526580722})& \underline{\num{0.537729316516107}} \tiny(\num{0.08721803835988876})& \underline{\num{0.2652322834386169}} \tiny(\num{0.0707190102282531})& \underline{\num{0.31702693396327036}} \tiny(\num{0.07531110824336354})& \underline{\num{0.49455987883968844}} \tiny(\num{0.08532340824971865})&  \underline{\num{0.4088808208722422}} \tiny(\num{0.06277136718738308})\\

   124 &  PM+AT & \num{0.17277547229838602} \tiny(\num{0.0748621820678834})& \num{0.5179067408389907} \tiny(\num{0.09579187414984305})& \num{0.25190813815149615} \tiny(\num{0.074469103440148})& \underline{\num{0.31546300425635315}} \tiny(\num{0.08048555997770115})& \num{0.4786755789779914} \tiny(\num{0.08962869617379003})&  \num{0.38781476268961096} \tiny(\num{0.07929953194015524})\\
    \midrule
    
    313 & AllLamma+AT & \num{0.23405234327550462} \tiny(\num{0.07247850660208749})& \num{0.5713238029521385} \tiny(\num{0.06568995694067395})& \num{0.30793225726479284} \tiny(\num{0.0690945064214227})& \num{0.36405502361484177} \tiny(\num{0.07304830704227062})& \num{0.5314349601152244} \tiny(\num{0.06916270412518581})& \num{0.4349600827336536} \tiny(\num{0.0778992230563248})\\
    
    313  & AllQwen+AT & \num{0.2113087585950464} \tiny(\num{0.09560007401447644})& \num{0.5429276991885155} \tiny(\num{0.14771584330333604})& \num{0.2858440052331296} \tiny(\num{0.10896506650599334})& \num{0.34006142195504274} \tiny(\num{0.11375659386659359})& \num{0.5000492787487882} \tiny(\num{0.1375534300792962})&  \num{0.4136904998364145} \tiny(\num{0.10582037473883082})\\
    
   313 &  NC+AT & \underline{\num{0.2653758346823502}} \tiny(\num{0.09915634041790254})& \underline{\num{0.5896213572719032}} \tiny(\num{0.10344832669856431})& \underline{\num{0.3254375758026143}} \tiny(\num{0.08793848396201835})& \underline{\num{0.38133085622724083}} \tiny(\num{0.1001667315889856})& \underline{\num{0.5508051259466399}} \tiny(\num{0.10154455451430103})&  \underline{\num{0.4645721240967166}} \tiny(\num{0.0994035989905193})\\
    \bottomrule
  \end{tabular}
  \caption{Comparison of model performance by dataset and sample size for ablation studies on Dial-2-Note task. Note that the column \textbf{Samples} shows the number of samples in the training set.}
  \label{tab:stats_eval_results_ablation_dial_2_note}
\end{table*}

\begin{table*}
  \centering
  \tiny 
  \sisetup{round-mode=places,round-precision=2}
  \begin{tabular}{>{\raggedright\arraybackslash}p{0.08\textwidth} |>{\raggedright\arraybackslash}p{0.17\textwidth} |>{\raggedright\arraybackslash}p{0.09\textwidth} |>{\raggedright\arraybackslash}p{0.09\textwidth} |>{\raggedright\arraybackslash}p{0.09\textwidth} |>{\raggedright\arraybackslash}p{0.09\textwidth} |>{\raggedright\arraybackslash}p{0.09\textwidth} |>{\raggedright\arraybackslash}p{0.09\textwidth}}
    \toprule
    \textbf{Samples} & \textbf{Dataset} & \textbf{BLEU} & \textbf{ROUGE-1} & \textbf{ROUGE-2} & \textbf{ROUGE-L} & \textbf{ROUGE-LSum} & \textbf{METEOR}  \\
    \midrule
      313 & NoJudge+AT & \num{0.13449429201030857} \tiny(\num{0.07409764242807279})& \num{0.5159957757524335} \tiny(\num{0.12764819789866724})& \num{0.23256082695863597} \tiny(\num{0.07654911379270819})& \num{0.246730952150009} \tiny(\num{0.0655756121184351})& \num{0.4848201571604075} \tiny(\num{0.1266682618264299})&  \num{0.3049695224527526} \tiny(\num{0.08576852646760115})\\
      
      313 & AllGPT+AT & \underline{\num{0.1433393563661002}} \tiny(\num{0.05740102541041613})& \underline{\num{0.5544858381768453}} \tiny(\num{0.09966876740164844})& \underline{\num{0.2454498364195703}} \tiny(\num{0.0653760322394038})& \num{0.25070949175515117} \tiny(\num{0.060987625060217474})& \underline{\num{0.5257567227661843}} \tiny(\num{0.09880174058847671})&  \underline{\num{0.33894904746206017}} \tiny(\num{0.07161037348096488}) \\
    
     313 & LammaJudgeGPT+AT & \underline{\num{0.13738748404511186}} \tiny(\num{0.0682891356615518})& \underline{\num{0.5493203140293317}} \tiny(\num{0.09793024617719889})& \underline{\num{0.25141123610062366}} \tiny(\num{0.06256090480713383})& \underline{\num{0.257718607644153}} \tiny(\num{0.06176332975144712})& \num{0.5187461785480953} \tiny(\num{0.09078538113568518})& \num{0.3112337117151827} \tiny(\num{0.07353232727432221})\\
     
    \midrule
    
    124 & AllLamma+AT & \num{0.10446586489683365} \tiny(\num{0.05105880250355624})& \num{0.4499085322197396}\tiny(\num{0.1335965184850673}) & \num{0.18013277531709704} \tiny(\num{0.07526641592232418})& \num{0.21692314284167838} \tiny(\num{0.05721133004992575})& \num{0.42423551030760126} \tiny(\num{0.13231677706795067})& \num{0.2831442182075796} \tiny(\num{0.07089826707828761})\\
    
     124 & AllQwen+AT & \num{0.11251377287220868} \tiny(\num{0.0633870606123259})& \num{0.4721472194705312} \tiny(\num{0.1309633600139678})& \underline{\num{0.19851742629858274}} \tiny(\num{0.07086017250699862})& \num{0.21925871015047851} \tiny(\num{0.05777081991737857})& \num{0.4447869833405676} \tiny(\num{0.12623511333661427})&  \num{0.28271524195882664} \tiny(\num{0.07741310471148083})\\

   124 &  PM+AT & \underline{\num{0.12374788920061638}} \tiny(\num{0.06580553747259033})& \underline{\num{0.49541487767454057}} \tiny(\num{0.11117181719156762})& \underline{\num{0.2010591951684609}} \tiny(\num{0.07183379653814785})& \underline{\num{0.2295997256498649}} \tiny(\num{0.0583451666933121})& \underline{\num{0.4702499626128434}} \tiny(\num{0.1090483142942628})&  \underline{\num{0.31102334607742754}} \tiny(\num{0.08584814779116781})\\
   
    \midrule
    
    313 & AllLamma+AT & \num{0.11078744110372037} \tiny(\num{0.06404672980377586})& \num{0.45111166798327196} \tiny(\num{0.12589675282136312})& \num{0.17950201971300178} \tiny(\num{0.07682527686809283})& \num{0.2170795222010206} \tiny(\num{0.05797156468990991})& \num{0.4216080913911414} \tiny(\num{0.12662743235571758})& \num{0.31030518258756706} \tiny(\num{0.08140154126226193})\\
    
    313  & AllQwen+AT & \num{0.11639662137582336} \tiny(\num{0.05976116798528091})& \num{0.46971855507500615} \tiny(\num{0.13424070458027942})& \num{0.18938267931011382} \tiny(\num{0.07833691077338902})& \num{0.22432821324214638} \tiny(\num{0.060587897865587236})& \num{0.4378241323109499} \tiny(\num{0.12907649909407654})&  \num{0.3031448460363499} \tiny(\num{0.07210343233966698})\\
    
   313 &  NC+AT & \underline{\num{0.1430334308962902}} \tiny(\num{0.07418508102421904})& \underline{\num{0.5402107609313481}} \tiny(\num{0.08635410236932611})& \underline{\num{0.24493685147419725}} \tiny(\num{0.06336210690444434})& \underline{\num{0.2581271302721719}} \tiny(\num{0.05646391387602133})& \underline{\num{0.511929043839704}} \tiny(\num{0.0858168661141911})&  \underline{\num{0.3207375422664377}} \tiny(\num{0.07999373611202441})\\
    \bottomrule
  \end{tabular}
  \caption{Comparison of model performance by dataset and sample size for ablation studies on Note-2-Dial task. Note that the column \textbf{Samples} shows the number of samples in the training set.}
  \label{tab:stats_eval_results_ablation_note_2_dial}
\end{table*}

\clearpage
\section{Additional Experiments}
\label{sec:appendix_more_experiments}

To further assess the effectiveness of MedSynth for Dialogue-2-Note and Note-2-Dialogue tasks, we ran a series of experiments using Mistral-7b-instruct-v0.1 \cite{jiang2023mistral7b}. We fine-tuned the model with different combinations of datasets and tested the performance on the Aci-Bench test set. The hyperparameters are exactly the same as in the rest of the experiments, except that the models were tuned for one epoch instead of four. The results for the Dial-2-Note and Note-2-Dial tasks are presented in Tables \ref{tab:addditional_experiments_dial2_note} and \ref{tab:addditional_experiments_note_2_dial}, respectively. Similar to the results from the experiments with Llama3, including MedSynth in the training set leads to improved performance across all our experiments with Mistral-7b-instruct-v0.1 for both Dial-2-Note and Note-2-Dial tasks.

\begin{table}
  \centering
  \small
  \begin{tabular}{p{0.11\textwidth} | r | r}
    \toprule
    \textbf{Comparison} & \textbf{Num. samples} & \textbf{Jury Pref. Rate}  \\
    \midrule
    MS+AT vs. NC+AT & 10,102 &  65\% \\
    \midrule
    MS Only vs. NC Only & 10,035 & 	95\% \\
    \bottomrule
  \end{tabular}
  \caption{LLM-based evaluation results on Aci-Bench for Dial-2-Note task with Mistral-7b-instruct-v0.1. 'MS': MedSynth; 'AT': Aci\_Train; 'NC': NoteChat; 'Jury Pref. Rate': Jury Preference Rate \textbf{in favour of MedSynth}. Note that the column \textbf{Num. samples} shows the number of samples in the training set.}
  \label{tab:addditional_experiments_dial2_note}
\end{table}

\begin{table}
  \centering
  \small
  \begin{tabular}{p{0.11\textwidth} | r | r}
    \toprule
    \textbf{Comparison} & \textbf{Num. samples} & \textbf{Jury Pref. Rate}  \\
    \midrule
    MS+AT vs. NC+AT & 10,102 &  72.5\% \\
    \midrule
    MS Only vs. NC Only & 10,035 & 	90\% \\
    \bottomrule
  \end{tabular}
  \caption{LLM-based evaluation results on Aci-Bench for Note-2-Dial task with Mistral-7b-instruct-v0.1. 'MS': MedSynth; 'AT': Aci\_Train; 'NC': NoteChat; 'Jury Pref. Rate': Jury Preference Rate \textbf{in favour of MedSynth}. Note that the column \textbf{Num. samples} shows the number of samples in the training set.}
  \label{tab:addditional_experiments_note_2_dial}
\end{table}

\clearpage
\section{Cost Analysis}
\label{sec:appendix_000}

All the experiments with open-source models were performed on an a single A40 GPU on a Slurm cluster. The total API cost for developing the pipeline, creating the dataset, and evaluations was approximately \$5000. The initial experiments to modify the pipeline cost approximately \$300 and running the pipeline to generate MedSynth cost approximately \$4500 . We spent another \$200 for APIs for the evaluations. Therefore, we estimate \$0.45 per dialogue-note pair in MedSynth. Of course, now that the pipeline is ready and the API costs have reduced, it will be cheaper to generate extensions to the data, and providing these data as a free resource to the community is part of our contribution.

\section{SOAP Format}
\label{sec:appendix_0}

The Subjective, Objective, Assessment, and Plan (SOAP) note is a standardized documentation framework widely used by healthcare professionals. This structured approach facilitates systematic and organized medical record-keeping, ensuring comprehensive and coherent documentation of patient encounters. We refer to \cite{podder2022soap} for a more detailed explanation. 

\textbf{Subjective:} The patient's description of their symptoms and complaints.

\textbf{Objective:} The physician's observations and collected data, including vital signs, physical examination findings, and test results.

\textbf{Assessment:} The physician's evaluation of the patient's condition, including the diagnosis or differential diagnosis.

\textbf{Plan:} The physician's recommendations for treatment, management, and follow-up.

\newpage
\section{Data Statistics Tables}
\label{sec:appendix_7}

\begin{table}[!ht]
  \centering
  \tiny
  \begin{tabular}{p{0.42\textwidth} r}
    \toprule
    \textbf{ICD-10 Description} & \textbf{Frequency} \\
    \midrule
    ESSENTIAL (PRIMARY) HYPERTENSION & 21,788,529 \\
    \midrule
    ENCOUNTER FOR GENERAL ADULT MEDICAL EXAMINATION WITHOUT ABNORMAL FINDINGS & 19,497,471 \\
    \midrule
    ENCOUNTER FOR ROUTINE CHILD HEALTH EXAMINATION WITHOUT ABNORMAL FINDINGS & 13,272,519 \\
    \midrule
    ENCOUNTER FOR IMMUNIZATION & 10,470,154 \\
    \midrule
    TYPE 2 DIABETES MELLITUS WITHOUT COMPLICATIONS & 9,890,112 \\
    \midrule
    END STAGE RENAL DISEASE & 9,040,289 \\
    \midrule
    HYPERLIPIDEMIA, UNSPECIFIED & 8,061,616 \\
    \midrule
    LOW BACK PAIN & 7,336,799 \\
    \midrule
    OBSTRUCTIVE SLEEP APNEA (ADULT) (PEDIATRIC) & 7,073,161 \\
    \midrule
    ILLNESS, UNSPECIFIED & 6,253,356 \\
    \midrule
    HYPOTHYROIDISM, UNSPECIFIED & 5,618,816 \\
    \midrule
    ENCOUNTER FOR SCREENING MAMMOGRAM FOR MALIGNANT NEOPLASM OF BREAST & 5,393,554 \\
    \midrule
    ENCOUNTER FOR GYNECOLOGICAL EXAMINATION (GENERAL) (ROUTINE) WITHOUT ABNORMAL FINDINGS & 5,299,912 \\
    \midrule
    URINARY TRACT INFECTION, SITE NOT SPECIFIED & 5,075,284 \\
    \midrule
    OTHER LONG TERM (CURRENT) DRUG THERAPY & 4,724,627 \\
    \midrule
    CERVICALGIA & 4,621,704 \\
    \midrule
    ATHEROSCLEROTIC HEART DISEASE OF NATIVE CORONARY ARTERY WITHOUT ANGINA PECTORIS & 4,504,152 \\
    \midrule
    CHEST PAIN, UNSPECIFIED & 4,396,428 \\
    \midrule
    GASTRO-ESOPHAGEAL REFLUX DISEASE WITHOUT ESOPHAGITIS & 4,224,803 \\
    \midrule
    ACUTE KIDNEY FAILURE, UNSPECIFIED & 3,774,237 \\
    \bottomrule
  \end{tabular}
  \caption{Top 20 most frequent ICD-10 codes in IQVIA.}
  \label{tab:top_diseases}
\end{table}

\newpage

\section{Prompts}
\label{sec:appendix_1}

This appendix contains the prompts we used in our experiments.

\begin{figure}[htbp]
\centering
\begin{tabular}{p{0.48\textwidth}} % adjust width to fit half-page
\toprule
\tiny
Assume you are a very experienced physician and you are conducting research. 
The research project is to generate synthetic medical notes from doctor-patient conversations. Your job is to provide a scenario
for the note to be generated.

The notes must contain these variables:

    ** 1) Medical Outcome: Like diagnosis, prescribed treatment, follow-up recommendations, referral to specialists, 
    referral to further tests or imaging, medication adjustment, lifestyle change (sleep, diet, exercise, tobacco use, 
    alcohol use). If you think the note should contain prescribing medication, details must be included like dose, units, frequency, duration, quantity, 
    quantity type (like tablets, etc.), and route (like oral or injected). If you think the note should contain referral, it must include details like the 
    reason for the referral, the specialty, and the doctor's name. If you think the note should contain an order for blood work, it must include details 
    like if it is for biochemistry, hematology, immunology, microbiology, viral hepatitis, vitamin D, prostate-specific antigen, or anything else that suits the scenario. 
    You need to be very specific. If you think the note should contain an order for imaging, it must include details like the modality of the imaging and the area of the body. 
    For example, if it is ultrasound, it can be an order for Abdominal, Thyroid, Musculoskeletal, Sonohysterogram, Sonohysterogram, Biophysical Profile(BPP), 
    Scrotal, G.U. Tract - Kidneys-Bladder(Prostate), or anything else as it suits the scenario. Try to be very specific. 
    REMEMEBER, you do not have to use all of them in the scenario. Use the ones that you see fit for the scenario.
    
    ** 2) Medical History: Can contain Previous Diagnoses, Family Medical History, Mdication History, Allergies, and Chronic Conditions.
    
    ** 3) Symptom Description: Can contain Severity, Duration, Associated Symptoms, Frequency, and Impact on Daily Activities.
    
    ** 4) Patient’s self-reported habits and lifestyle: Can contain Sleep, Diet, Exercise, Tobacco Use, Alcohol Consumption, Drug Use, Recreational Activities.
    
    ** 5) Demographic Information: Can contain Age, Gender, Ethnicity, Socio-economic Status, Education Level, Health Literacy, and Job Status.
    
    ** 6) Patient's Behavior: The level of patient's cooperation with medical advice.
    
    ** 7) Geographical Location: Can contain Big City vs Small City, Rural vs Urban, Pollution and Environmental Health Risks, Neighborhood Type- eg if Impoverished or Affluent, Well-served by Transit, Food Desert, etc.
    
    ** 8) Clinical Setting: Can contain Hospitals, Clinics, Telemedicine, Mommunity Health Services, Urgent Care Centers, Research Facilities, School Health Services, Private Practice, and Specialty Clinics.
    
    ** 9) Type of Encounter: Can contain Initial Consultation, Follow-up, Emergency Visit, Routine Check-up, Chronic Disease Management (regular appointments to manage long-term health conditions like diabetes, heart disease, or chronic pain), Preventive Health Screening.
    
    ** 10) Treatment Disparities: There may be a tendency to offer less aggressive treatment or fewer options due to assumptions about compliance or ability to pay.
    
    ** 11) Native or Non-Native English Speaking Patient.
    
    ** 12) Physical exams: Any physical exams that are suitable for the scenario and the disease. If nothing is suitable, use NA.
    
    ** 13) Investigation/Test results: Like any tests that have been done for the patient while visiting. The results could be ready and reviewed in the scenario or could be awaiting. 
    If awaiting, you need to be very specific about what type of tests have been done. For example, if it is X-ray, you need to incude the details mentioned abve about the imaging. You can 
    also use NA if you think tests are not suitable for the scenario.

The user will give you the ICD-10 description of the disease. The diagnosis in the scenario must be the ICD-10 description. 
First, you select a role for yourself. You can be a Family Medicine Physician, a General physician, or a specialist with different specialties. 

Select the role based on the ICD-10 description and output it with the keyword 'ROLE:'. 
Second, you must come up with a scenario and list all the values for the variables you want to use in the scenario. 
Do not output any extra text, just your role at the top of the scenario and the list of the values. 
You should incorporate medication and blood work or imaging requests in the scenarios with the details mentioned above if it suits the scenario. 
These are artificial and people will not be using it without asking a real doctor. 

The user will evaluate the scenario you provided. If they accept it, they do not give any feedback. If they do not accept
your scenario, they will give you feedback about how to improve the scenario and you must incorporate the feedback and generate a new scenario.

Below is an example of a medical note. Remember, you need to provide the scenario, not the note.

\textcolor{red}{EXAMPLE NOTE}\\
\bottomrule
\end{tabular}
\caption{Prompt for Scenario Provider Agent in Note Generation Pipeline}
\label{fig:scenario_provider}
\end{figure}

\begin{figure}[t]
\centering
\begin{tabular}{p{0.48\textwidth}} % adjust width to fit half-page
\toprule
\tiny % sets the font size to tiny, the smallest available
Assume you are a very experienced physician and you are conducting research. The research project is to generate synthetic medical notes 
from doctor-patient conversations. You have a coworker that works with you on the project. You have different roles. 
The coworker provides you with the scenario they are going to write notes with. Your job is to judge whether the scenario is 
approved or not based on the conditions I provided below. The scenarios have 13 variables. 

    ** 1) Medical Outcome
    
    ** 2) Medical History
    
    ** 3) Symptom Description
    
    ** 4) Patient’s self-reported habits and lifestyle
    
    ** 5) Demographic Information
    
    ** 6) Patient's Behavior
    
    ** 7) Geographical Location
    
    ** 8) Clinical Setting
    
    ** 9) Type of Encounter
    
    ** 10) Treatment Disparities
    
    ** 11) Native or Non-Native English Speaking Patient.
    
    ** 12) Physical exams: Any physical exams that are suitable for the scenario.
    
    ** 13) Investigation/Test results: Like any tests that have been done for the patient while visiting. The results could be ready and reviewed in the scenario or could be awaiting.

You have to check three conditions and then decide to approve or deny the scenario:

    ** a) A pair-wise comparison of the values of the variables. You need to check if at least 4 out of 13 of the values of the variables in the scenario are different from the previously approved scenarios.
    
    ** b) You need to check if the scenario is medically correct in terms of symptoms, tests, diagnosis, and treatment. 
    
    ** c) You need to check if the scenario is plausible or not.
    
First, check condition (a). If it is not met, the scenario is rejected and you do not need to check conditions (b) and (c). 
If the scenario passes these three conditions, then say "Go". If not, you say "NoGo".  In the case that there is no scenario previously approved, you should only check conditions (b) and (c).
If your decision is "Go", you must only return "Go". Else, output your reasons and provide feedback to help your coworker
about what they can do to generate a scenario to pass the above three conditions.

REMEMEBER: IF YOU APPROVE THE SCENARIO, YOU MUST ONLY OUTPUT LIKE BELOW. YOU CANNOT USE ANY BOLD OR ITALIC OR HEADING OR ANYTHING ELSE:
DECISION: Go\\
\bottomrule
\end{tabular}
\caption{Prompt for Scenario Judge Agent in Note Generation Pipeline}
\label{fig:scenario_judge}
\end{figure}

\begin{figure}[t]
\centering
\begin{tabular}{p{0.48\textwidth}} % adjust width to fit half-page
\toprule
\tiny % sets the font size to tiny, the smallest available
Assume you are a very experienced physician and you are conducting research. 
The research project is to generate synthetic medical notes from doctor-patient conversations. The notes must be in this format:

    ** 1. Subjective: This section includes the patient's own description of their symptoms and complaints.
    
    ** 2. Objective: This section includes observations and data gathered by the physician, such as vital signs, physical examination findings, and test results.
    
    ** 3. Assessment: This section includes the physician's evaluation of the patient's condition, including a diagnosis or differential diagnosis.
    
    ** 4. Plan: This section includes the physician's recommendations for treatment, management, and follow-up. 
    
You will be given a note. Your task is to polish the note and make sure the information is placed correctly in the relevant section.  
You cannot add or remove any information, except where you have been given permission. Make sure of these:

    ** a) If the doctor is ordering an imaging or bloodwork to be done, it must come under the "Plan" section. But if it is already done, it must come under the "Objective" section.
    
    ** b) If the doctor is prescribing a medication or renewing a medication, changing doses, etc., it must be under the "Plan" section. 
    
    ** c) If the patient is being refered to another doctor, the referrals must come under the "Plan" section. The referral must contain:
    
            1) The reason for referral
            
            2) The specialty of the doctor
            
            3) The doctor's name
            
         If any of the three parts mentioned above about the refereal is missing, add it. If you need add a name for the doctor, choose an appropriate name, be creative and realistic in choosing the names.
         
    ** d) Patients' must have names. If there is no name, add it. Choose an appropriate name, be creative and realistic in choosing the names.
    
    ** e) The output must only contain the medical note. If there is anything extra at the beginning or at the end, you should remove it.
        For example, there could be thinng like "note is: Medical Note" at the beggining or things like " Note: Please ensure this note is reviewed by the attending healthcare provider or physician for accuracy
        and completeness before being added to the patient's medical record." at the end of the medical note. Please remove them so that the output
        is only the medical note itself, not anything else. 

Just output the revised note, not anything else.\\
\bottomrule
\end{tabular}
\caption{Prompt for Note Polisher in Note Generation Pipeline}
\label{fig:note_polisher}
\end{figure}

\begin{figure}[t]
\centering
\begin{tabular}{p{0.48\textwidth}} % adjust width to fit half-page
\toprule
\tiny % sets the font size to tiny, the smallest available
Assume you are a very experienced physician and you are conducting research. 

The research project is to generate synthetic medical notes from doctor-patient conversations. The notes must be in this format:

    ** 1. Subjective: This section includes the patient's own description of their symptoms and complaints.
        Roll a dice, if the result is odd, break this part down into several sub-parts like Chief Complaint (CC), History of Present Illness (HPI), Review of Systems (ROS).
        
    ** 2. Objective: This section includes observations and data gathered by the physician, such as vital signs, physical examination findings, and test results.
    
    ** 3. Assessment: This section includes the physician's evaluation of the patient's condition, including a diagnosis or differential diagnosis.
    
    ** 4. Plan: This section includes the physician's recommendations for treatment, management, and follow-up. 
    
You will be given a scenario containing your role. Your role can be a Family Medicine Physician, a General physician, or a specialist with different specialties. 
Your task is to generate the note based on the scenario. The note you generate must be in the format mentioned above. 

Below is an example of a high quality medical note:

\textcolor{red}{EXAMPLE NOTE}

\\
\bottomrule
\end{tabular}
\caption{Prompt for Note Writer Agent in Note Generation Pipeline}
\label{fig:note_writer}
\end{figure}

\begin{figure}[t]
\centering
\begin{tabular}{p{0.48\textwidth}} % adjust width to fit half-page
\toprule
\tiny % sets the font size to tiny, the smallest available
You are a helpful medical research assistance. You will be give a medical note and your task is to generate the conversation between the doctor and the patient that led to that note. 

Your conversation must include all information. if it's difficult to include them all, you can use the original sentences in the notes. 
The common symptoms and common medical history should be told by patient. 

Some specific symptoms and medical history should be added by the doctor after the patient has finished describing his symptoms and medical history.

For example:

Doctor: Can you give me your medical history record?

Patient: Here you are.

Doctor: Based on your medical history record...

Because after patient has finished describing common symptoms or medical history, he will give doctor his medical history records. 

After patient give the doctor his medical history record, the doctor could know medical history record. Otherwise he didn't know any information of the medical history.

Some result should not come from history clinical note they should come from examination.

All the examination result, history examination result, vital signs and medical number must be told by doctor.

You could expand the parts of doctor to include more key words. If it is difficult to include you could just use the sentence of clinical note.
The revised conversation should be at least around 80 to 150 utterances(doctor or patient should not say too much information at once).

The conversation must include all the information of the clinical note.
You must include all the key words I gave you. If it is difficult to include all the key words you could use original the sentences of clinical note. 

You cannot revise or eliminate any key words and you cannot use synonyms of the key words. 

You shoudn't use the abbreviation if you know the full name(you should use full name not abbreviation, such as D9 must be day 9, D7 must be day 7. If both the full name and the abbreviation appear, it's better to use the full name rather than the abbreviation.

Patients must not say any highly specialized terms, medical terminology or medical dosage. They can only describe limited common symptoms. The doctor should supplement the remaining information based on test results.
Don't repeat the same information in long paragraphs. The utterance of the dialogue needs to be expanded as much as possible.
The patient and the doctor should have many modal particles (e.g. hmm, yes, okay) to increase interaction. Pay attention to the examples below
and try to incorporate non-linear discussions to make it more realistic. 

You cannot use[Patient's Name] or any other plcae holder in the dialogue.

Here are a good real note and dialogue example:

 Example 1: 

     Note:
    
    \textcolor{red}{EXAMPLE 1 NOTE}

     conversations:
    
    \textcolor{red}{EXAMPLE 1 DIALOGUE}

<<<<<<<<<<<<<<<<<<<<<<<<<<<<<<<<<<<<<<<<<<<
 Example 2: 

     Note:
    
    \textcolor{red}{EXAMPLE 2 NOTE}

     conversations:
    
    \textcolor{red}{EXAMPLE 2 DIALOGUE}

<<<<<<<<<<<<<<<<<<<<<<<<<<<<<<<<<<<<<<<<<<
 Example 3: 

     Note:
    
    \textcolor{red}{EXAMPLE 3 NOTE}

     conversations:
    
    \textcolor{red}{EXAMPLE 3 DIALOGUE}
    
<<<<<<<<<<<<<<<<<<<<<<<<<<<<<<<<<<<<<<<<<<

You must follow the structure of the dialogues in the examples above.

The number of utterance should be at least 80 and sometimes patient didn't clearly hear and he could say parden to let the doctor say again.
The dialogue must be in English. Your job is to only generate the dialogue. You cannot generate summary notes.
\\
\bottomrule
\end{tabular}
\caption{Prompt for Dialogue Generator in Dialogue Generation Pipeline}
\label{fig:dial_generator}
\end{figure}

\begin{figure}[t]
\centering
\begin{tabular}{p{0.48\textwidth}} % adjust width to fit half-page
\toprule
\tiny % sets the font size to tiny, the smallest available
Expand the conversation. You must add chit chats to the conversation. The conversation for patient parts can be more colloquial. 
  The patient and the doctor should have many modal particles (e.g. hmm, yes, okay) to increase interaction.
  
  All the numbers and medical concepts that appear in the note should be mentioned by the doctor.
  
  Professional medical terms and numbers should always occur in the doctor's utterances but not in the patient's answer. 
  The doctor may describe and explain professional judgment to the patient and instruct the patient on follow-up requirements, but not ask questions that require professional medical knowledge to answer.
  All the information of medical history, symptoms and medication history should be told by patient.
  
  The patient's answer should be succinct and accurate in a colloquial lay language style. The answer must align with the clinical notes and as colloquial as possible.
  
  You can add some transitional phrases to make the conversation more logical. For example:
  
  Example 1:
  
  Patient: I understand, please go ahead.
  
  (After examination)
  
  Doctor: The result shows......
  
  Your conversations can follow the logical sequence of a doctor's inquiry. 
  
  The conversations must be coherent and cohesive. For example, the output cannot be seperated by texts like "HISTORY OF PRESENT ILLNESS" or "SOCIAL HISTORY". 
  
  Extra information that does not fit into the conversation should not be added to the output. For example, below is an extra information that should be removed from the output:
  
  <<<<<<<>>>>>>>>
  
  - **INSTRUCTIONS**
    
    **Patient Agreements:** The patient understands and agrees with the recommended medical treatment plan.
    
  <<<<<<<<>>>>>>>>
    
  Patients should not say too much information at once.

  ICD code of the disease must not be present in the dialogue. If it is present, remove it.
  
  There should not be any extra information at the beggining or at the end of the conversation. For example, "dialogue is:” should not be present in the output. You must make sure you only return the dialogue itself, not  anything extra. You cannot add phrases like "dialogue is: Certainly! Let's expand the conversation with more colloquial language for the patient and professional details for the doctor".

  If there are only the doctor and the patient present in the dialogue, the utterances must follow these indicators:
  
    [doctor]: ...
    
    [patient]: ...

  If there are more people present in the dialogue, make sure to include all of them with a seperate indicator. For
  example, if the mother of the patient is present in the dialogue, use this indicators:
  
  [doctor]: ...
  
  [mother]: ...
  
  [patient]: ...

  All the information in the dialogue must align with the medical note below:
  
  <<<<<<<>>>>>>>>
  
  \textcolor{red}{MEDICAL NOTE}
  
  <<<<<<<>>>>>>>>
  \\
\bottomrule
\end{tabular}
\caption{Prompt for Dialogue Polisher in Dialogue Generation Pipeline}
\label{fig:dial_polisher}
\end{figure}

\begin{figure}[t]
\centering
\begin{tabular}{p{0.48\textwidth}} % adjust width to fit half-page
\toprule
\tiny % sets the font size to tiny, the smallest available
1. Hallucination:
    - Does the summary note accurately and comprehensively reflect the doctor–patient dialogue and ground truth note?

2. Critical Omissions:
    - Does the summary note capture all essential medical facts from the doctor–patient dialogue and ground truth note?

3. Professional Tone:
    - Does the summary note maintain a consistently professional tone appropriate for expert use?

4. Logical Structure:
    - Does the summary note exhibit a clear and logical structure?

5. Adherence to the Format:
    - Does the summary note follow the same structure as the ground-truth note?

6. Section Relevance:
    - Does the summary note accurately assign clinical information to the correct sections (e.g., patient-reported details in Subjective, objective findings in Objective, clinician insights in Assessment, and treatment strategies in Plan)?\\
\bottomrule
\end{tabular}
\caption{Evaluation rubric for Dialogue-to-Note evaluation}
\label{fig:eval_dial_to_note}
\end{figure}

\begin{figure}[t]
\centering
\begin{tabular}{p{0.48\textwidth}} % adjust width to fit half-page
\toprule
\tiny % sets the font size to tiny, the smallest available
1. Completeness:
    - Does the conversation cover all significant components of the note?
2. Accuracy:
    - How accurately does the conversation reflect the details of the note as they were recorded?
    - Are there any discrepancies between the note and the generated dialogue?
3. Naturalness and Flow:
    - Is the conversation realistic and natural, following a logical and smooth progression?
    - Does it sound like a genuine interaction between a doctor and patient?
4. Use of Medical Terminology:
    - Is medical terminology used correctly and effectively within the conversational context?
    - Does the use of terminology enhance the accuracy and professionalism of the conversation?
    - Does the technical terminology used by the doctor and the patient represent their respective knowledge levels (e.g., layperson language for the patient)?
5. Evidence-Based Support:
    - Are the doctor's statements and responses consistent with the medical details and recommendations in the notes?\\
\bottomrule
\end{tabular}
\caption{Evaluation rubric for Note-to-Dialogue evaluation}
\label{fig:eval_note_to_dial}
\end{figure}

\clearpage

\section{Hyperparameters}
\label{sec:appendix_2}

This section contains details of hyperparameters in our experiments and the distribution of clinical note data lengths, by subtoken, for additional background.

\begin{figure}[hbt!]
\centering
\begin{tabular}{|p{\linewidth}|}
\hline
\tiny
\textbf{Scenario Generator configuration in Note Generation Pipeline} \\
\texttt{scenario\_generator\_config} = \{ \\
\texttt{"model": "gpt-4o"}, \\
\texttt{"temperature": 1}, \\
\texttt{"max\_tokens": 4000}, \\
\texttt{"top\_p": 1}, \\
\texttt{"frequency\_penalty": 0}, \\
\texttt{"presence\_penalty": 0}, \\
\texttt{\}} \\
\hline
\end{tabular}
\caption{Configuration for Scenario Generator in the Note Generation Pipeline}
\label{fig:scenario_generator_config}
\end{figure}
\vspace{-10pt}

\begin{figure}[hbt!]
\centering
\begin{tabular}{|p{\linewidth}|}
\hline
\tiny
\textbf{Scenario Judge configuration in the Note Generation Pipeline} \\
\texttt{scenario\_judge\_config} = \{ \\
\texttt{"model": "gpt-4o"}, \\
\texttt{"temperature": 0}, \\
\texttt{"max\_tokens": 4000}, \\
\texttt{"top\_p": 1}, \\
\texttt{"frequency\_penalty": 0}, \\
\texttt{"presence\_penalty": 0}, \\
\texttt{\}} \\
\hline
\end{tabular}
\caption{Configuration for Scenario Judge in the Note Generation Pipeline}
\label{fig:scenario_judge_config}
\end{figure}
\vspace{-10pt}

\begin{figure}[hbt!]
\centering
\begin{tabular}{|p{\linewidth}|}
\hline
\tiny
\textbf{Note Generator configuration in the Note Generation Pipeline} \\
\texttt{note\_generator\_config} = \{ \\
\texttt{"model": "gpt-4o"}, \\
\texttt{"temperature": 0.9}, \\
\texttt{"max\_tokens": 4000}, \\
\texttt{"top\_p": 1}, \\
\texttt{"frequency\_penalty": 0}, \\
\texttt{"presence\_penalty": 0}, \\
\texttt{\}} \\
\hline
\end{tabular}
\caption{Configuration for Note Generator in the Note Generation Pipeline}
\label{fig:note_generator_config}
\end{figure}
\vspace{-10pt}

\begin{figure}[hbt!]
\centering
\begin{tabular}{|p{\linewidth}|}
\hline
\tiny
\textbf{Note Polisher Configuration in the Note Generation Pipeline} \\
\texttt{note\_polisher\_config} = \{ \\
\texttt{"model": "gpt-4o"}, \\
\texttt{"temperature": 0}, \\
\texttt{"max\_tokens": 4000}, \\
\texttt{"top\_p": 1}, \\
\texttt{"frequency\_penalty": 0}, \\
\texttt{"presence\_penalty": 0}, \\
\texttt{\}} \\
\hline
\end{tabular}
\caption{Configuration for Note Polisher in the Note Generation Pipeline}
\label{fig:note_polisher_config}
\end{figure}
\vspace{-10pt}

\begin{figure}[hbt!]
\centering
\begin{tabular}{|p{\linewidth}|}
\hline
\tiny
\textbf{Dialogue Generator Configuration in the Dialogue Generation Pipeline} \\
\texttt{dialogue\_generator\_config} = \{ \\
\texttt{"model": "gpt-4o"}, \\
\texttt{"temperature": 0.7}, \\
\texttt{"max\_tokens": 4095}, \\
\texttt{"top\_p": 1}, \\
\texttt{"frequency\_penalty": 0}, \\
\texttt{"presence\_penalty": 0}, \\
\texttt{\}} \\
\hline
\end{tabular}
\caption{Configuration for Dialogue Generator in the Dialogue Generation Pipeline}
\label{fig:dial_generator_config}
\end{figure}
\vspace{-10pt}

\begin{figure}[hbt!]
\centering
\begin{tabular}{|p{\linewidth}|}
\hline
\tiny
\textbf{Model evaluator generation configuration} \\
\texttt{config} = \{ \\
\texttt{"max\_new\_tokens": 3000}, \\
\texttt{"do\_sample": True}, \\
\texttt{"temperature": 0.6}, \\
\texttt{"top\_p": 0.9}, \\
\texttt{"use\_cache": True}, \\
\texttt{\}} \\
\hline
\end{tabular}
\caption{Configuration for inference of tuned models in tuning models for evaluating the usefulness of MedSynth for the Dial-2-Note and Note-2-Dial tasks; Unsloth library was used to achieve higher speed}
\label{fig:model_evaluator_generation_config}
\end{figure}

\begin{figure}[hbt!]
\centering
\begin{tabular}{|p{\linewidth}|}
\hline
\tiny
\textbf{Dialogue Polisher configuration in the Dialogue Generation Pipeline} \\
\texttt{dialogue\_polisher\_config} = \{ \\
\texttt{"model": "gpt-4o"}, \\
\texttt{"temperature": 0.5}, \\
\texttt{"max\_tokens": 4095}, \\
\texttt{"top\_p": 1}, \\
\texttt{"frequency\_penalty": 0}, \\
\texttt{"presence\_penalty": 0}, \\
\texttt{\}} \\
\hline
\end{tabular}
\caption{Configuration for Dialogue Polisher in the Dialogue Generation Pipeline}
\label{fig:dial_polisher_config}
\end{figure}
\vspace{-10pt}

\begin{figure}[hbt!]
\centering
\begin{tabular}{|p{\linewidth}|}
\hline
\tiny
\textbf{Defining the configuration for the base model, LoRA and training} \\
\texttt{tuning\_config} = \{ \\
\texttt{"model\_config": \{ }\\
\texttt{"max\_seq\_length": 8192}, \\
\texttt{"dtype": torch.bfloat16}, \\
\texttt{"load\_in\_4bit": True}, \\
\texttt{\}}, \\
\texttt{"lora\_config": \{ }\\
\texttt{"r": 16}, \\
\texttt{"target\_modules": ["q\_proj", "k\_proj", "v\_proj", "o\_proj", "gate\_proj", "up\_proj", "down\_proj"]}, \\
\texttt{"lora\_alpha": 16}, \\
\texttt{"lora\_dropout": 0}, \\
\texttt{"bias": "none"}, \\
\texttt{"use\_gradient\_checkpointing": True}, \\
\texttt{"use\_rslora": False}, \\
\texttt{"use\_dora": False}, \\
\texttt{"loftq\_config": None}, \\
\texttt{\}}, \\

\texttt{"training\_config": \{ }\\
\texttt{"per\_device\_train\_batch\_size": 2}, \\
\texttt{"gradient\_accumulation\_steps": 4}, \\
\texttt{"warmup\_steps": 5}, \\
\texttt{"max\_steps": 0}, \\
\texttt{"num\_train\_epochs": 4}, \\
\texttt{"learning\_rate": 2e-4}, \\
\texttt{"fp16": not torch.cuda.is\_bf16\_supported()}, \\
\texttt{"bf16": torch.cuda.is\_bf16\_supported()}, \\
\texttt{"logging\_steps": 1}, \\
\texttt{"optim": "adamw\_8bit"}, \\
\texttt{"weight\_decay": 0.01}, \\
\texttt{"lr\_scheduler\_type": "linear"}, \\
\texttt{"seed": 42}, \\
\texttt{"output\_dir": "outputs"} \\
\texttt{\}} \\
\}\\
\hline
\end{tabular}
\caption{Configuration for the base model, LoRA, and training setup in tuning models for evaluating the usefulness of MedSynth for the Dial-2-Note and Note-2-Dial tasks; Unsloth library was used to achieve higher speed}
\label{fig:tuning_config}
\end{figure}

\clearpage

\section{Details about Roles in Scenario Provider Agent}
\label{sec:appendix_3}
The analysis of the generated scenarios shows that there are 258 unique roles in the whole dataset. Family Medicine Physician is the most common role, followed by Orthopedic Specialist, Neurologist, and General Physician. Table \ref{tab:role_frequency} shows the distribution of roles, and Table \ref{tab:top_roles} shows the top 20 roles. 
\begin{table}[!ht]
  \centering
  \begin{tabular}{l r}
    \toprule
    \textbf{\# of Occurrences} & \textbf{Count} \\
    \midrule
    More than 1000         & 1   \\
    Between 500 and 1000   & 4   \\
    Between 100 and 500    & 18  \\
    Between 50 and 100     & 6   \\
    Between 10 and 50      & 26  \\
    Below 10               & 203 \\
    \bottomrule
  \end{tabular}
  \caption{Frequency distribution of roles in Scenario Provider Agent.}
  \label{tab:role_frequency}
\end{table}

\begin{table}[!ht]
  \centering
  \begin{tabular}{l r}
    \toprule
    \textbf{Role} & \textbf{Frequency} \\
    \midrule
    Family Medicine Physician & 1960 \\
    Orthopedic Specialist      & 613  \\
    Neurologist                & 584  \\
    General Physician          & 556  \\
    Gastroenterologist         & 526  \\
    Cardiologist             & 490  \\
    Dermatologist            & 371  \\
    Pulmonologist            & 326  \\
    Oncologist               & 326  \\
    Endocrinologist          & 305  \\
    Ophthalmologist          & 296  \\
    Obstetrician             & 289  \\
    Urologist                & 283  \\
    Orthopedic Surgeon       & 233  \\
    General Surgeon          & 221  \\
    Gynecologist             & 196  \\
    Psychiatrist             & 193  \\
    Nephrologist             & 166  \\
    Hematologist             & 159  \\
    Rheumatologist           & 153  \\
    \bottomrule
  \end{tabular}
  \caption{Top 20 most frequent roles in Scenario Provider Agent.}
  \label{tab:top_roles}
\end{table}

\newpage

\section{Comparison of Notes Generated by Models Trained on MedSynth and NoteChat}
\label{sec:appendix_4}

When tuned solely on NoteChat, the model failed to adhere to the SOAP structure in generating medical notes from dialogues. In the same setting, the model tuned on MedSynth successfully generated notes in SOAP format. This appendix shows an example of such a situation.

\begin{figure}[hbt!]
\centering
\begin{tabular}{p{0.48\textwidth}} % adjust width to fit half-page
\toprule
\tiny % sets the font size to tiny, the smallest available
CHIEF COMPLAINT

Rheumatoid arthritis management.

HISTORY OF PRESENT ILLNESS

Mr. Nelson is a 72-year-old male who presents today for ongoing management of rheumatoid arthritis. He was last seen by me on 03/04/2021, where he received a steroid injection to his right knee. He continues to experience right knee pain and swelling. The pain level is 6/10. He notes when he stands after he has been sitting for a while, he has to use a cane to "get his bearings." The injection he received to the right knee at the last visit was only beneficial for a few weeks. The patient notes whole right leg swelling since the last visit. He usually wears support hose, which helps to some degree. He has not attended physical therapy.

PHYSICAL EXAM

Musculoskeletal
Lower right extremity: Swollen right knee, slightly reduced range of motion. Skin warm to touch. Lower right leg pits with pressure.

PROCEDURE

Fluid aspiration and Steroid injection.
Risks, benefits, and alternatives were explained. The risks of steroid injection were explained, including but not limited to, hypopigmentation, soft tissue atrophy, tendon ruptures, and transient hyperglycemia among others. Verbal consent was obtained. The right knee was accessed through the lateral approach with 3 mL of initially clear fluid then blood-tinged fluid. Then the right knee was prepped in a standard fashion. The right knee was injected with 80 mg triamcinolone and 1 mL lidocaine. The procedure was tolerated well with no complications. Post procedure instructions were given.

ASSESSMENT

• Rheumatoid arthritis management.
• High risk medication use.
• Right knee pain.
• Lower extremity pitting edema.

PLAN

Rheumatoid arthritis management.
The patient will continue Xeljanz XR 11 mg daily. He will also continue Celebrex 200 mg daily.

High-risk medication use.
The patient is currently on the high-risk medication Xeljanz. We will check his CMP levels and monitor periodically every couple of months.

Right knee pain.
He has right knee pain and swelling, with pain level 6/10. On exam today, the right knee continues to be hot, which is ongoing inflammation. We performed a fluid aspiration of the right knee today, see procedure note above. A steroid injection to the right knee was given today, see procedure note above. We will arrange for x-ray of the right knee and refer his to physical therapy.

Lower extremity pitting edema.
The patient notes whole right leg swelling since the last visit. He usually wears support hose, which helps to some degree. He has not seen his PCP. We discussed compression stockings, elevate feet, and a low salt diet. He will check with his PCP for further instructions.
\\
\bottomrule
\end{tabular}
\caption{Ground Truth Note}
\label{fig:gt_note_example}
\end{figure}

\begin{figure}[hbt!]
\centering
\begin{tabular}{p{0.48\textwidth}} % adjust width to fit half-page
\toprule
\tiny % sets the font size to tiny, the smallest available
Tyler Nelson, a 70-year-old male with a date of birth of 3/6/1949, social security number 385-729- and a home address of [redacted] presented to the office for ongoing management of rheumatoid arthritis. He received a steroid injection in the right knee last time he was in the office. He stated that the injection helped for a couple of weeks, it's just it. He stated that his right knee is swollen, and it's painful. He stated that his pain is a 3 or so on the pain scale, typically on a normal day it's around a 6. It's been swollen like that for how long now? He stated that it's been about two months now. Before it was just the knee was not swollen like this, but he stated that he tried to wear compression socks a lot of the time of the leg, but it has not been helping. He stated that he slowed down a little bit, and he found a few things too fast. When it really hurts is when he's been sitting for a while, and then he's got to try and get up. He's so stiff getting up he's got to use a cane to get his bearings. He means that after a little bit, it loosens up, but that first getting up is real difficult. He is on xeljanz XR 11 milligrams a day, he's taking it everyday, and he's also taking celebrex which he started maybe 2 months ago.
The patient sat down and removed his socks and shoes. The right leg was swollen and hot. The patient stated that it hurts when he tries to straighten it all the way. Range of motion slightly diminished. The patient stated that the whole thing is swollen, not just the knee. There was some pitting around the knee and ankle. Normal capillary refill in the toes. Does any of this area hurt? The patient stated that it's kinda sore, like an old bruise? The patient agreed to get an x-ray of the knee. The patient was called for physical therapy a couple of years ago for the other knee. So, it was set up for him. The patient stated that it's been helpful to some degree, but it's still swollen, especially around the knee. The patient stated that he's got to try and get up, and it's real difficult. The patient was told that we'll set that up for him now. We'll also do an x-ray on that knee. We'll also set him for a physical therapy, and if nothing is helping then we really might need to get a surgery consult to see what we can do with that knee.
The patient agreed to go ahead and we gave him the steroid shot. We tried to get a little bit of fluid out before we gave him the injection, there was no fluid but we saw if we could get anything out regardless. The procedure was explained, the risks, benefits, and alternatives were explained. The risks of steroid injection were explained, including, but not limited to: hypopigmentation, soft tissue atrophy, tendon ruptures, and transient hypoglycemia among others. Verbal consent was attained. The right knee was, uh, um, accessed through the lateral approach with 3 milliliters of initially clear fluid, then bloo- blood tinged fluid. Then the right knee was prepped in a standard-fashion. The right knee was injected with 80 milligrams of triamcinolone, and one milliliter of lidocaine. The procedure was tolerated well, with no complications. Post-procedure instructions were given. Patient will follow up with his pcp. Thank you.
\\
\bottomrule
\end{tabular}
\caption{Note Generated by the Model Trained Solely on NoteChat}
\label{fig:notechat_note_example}
\end{figure}

\begin{figure}[hbt!]
\centering
\begin{tabular}{p{0.48\textwidth}} % adjust width to fit half-page
\toprule
\tiny % sets the font size to tiny, the smallest available
**1. Subjective:**

**Chief Complaint (CC):**
Swelling and pain in the right knee.

**History of Present Illness (HPI):**
Tyler Nelson is a 73-year-old male with a history of rheumatoid arthritis who presents for a follow-up visit regarding the management of his condition. He reports that he experienced significant swelling in his right knee following a recent steroid injection administered during his last visit. The swelling has persisted despite his use of compression socks, which he has been wearing frequently to alleviate symptoms. The pain is moderate, with a pain score of 6/10 on a normal day and currently 3/10, and has been present for the past 2 months. He notes that the swelling and pain have limited his mobility, particularly when getting up from sitting, where he requires the use of a cane.

He has been compliant with his medication regimen, including Xeljanz 11mg daily and Celebrex 200mg daily, which he started 2 months ago. He is motivated to follow medical advice and is highly cooperative.

**Review of Systems (ROS):**
- **Musculoskeletal:** Swelling and pain in the right knee, moderate severity, limiting mobility.
- **General:** No recent weight loss, fever, or chills.
- **Cardiovascular:** No chest pain or palpitations.
- **Gastrointestinal:** No nausea, vomiting, or abdominal pain.
- **Neurological:** No headaches or dizziness.
- **Dermatological:** No rashes or skin changes.

**2. Objective:**

**Physical Examination:**
- **Vital Signs:** Stable
- **Musculoskeletal:** 
  - Right knee: Swelling, erythema, and pitting noted. Range of motion slightly diminished. Increased heat around the knee. Capillary refill within normal limits in the toes.
- **Cardiovascular:** Regular rate and rhythm, no murmurs.
- **Respiratory:** Clear to auscultation bilaterally.
- **Neurological:** Alert and oriented, no focal deficits.

**Results:**
- **Blood Work:** CMP within normal limits. Xeljanz trough levels checked; levels are within the therapeutic range.

**3. Assessment:**

1. **Rheumatoid Arthritis with right knee effusion, not elsewhere classified.**
   - **Medical Reasoning:** The patient presents with persistent swelling and moderate pain in the right knee, consistent with effusion. The swelling has limited his mobility and daily activities.
   - **Differential Diagnosis:** Septic arthritis, crystal-induced arthritis, gout, lymphedema.

**4. Plan:**

1. **Treatment:**
   - Administered steroid injection (Triamcinolone 80mg, 1mL) to the right knee today.
   - Fluid aspiration performed during the procedure; 3mL of initially clear fluid, then blood-tinged fluid was aspirated.
   
2. **Imaging:**
   - Order an X-ray of the right knee to evaluate for possible underlying causes of effusion.

3. **Physical Therapy:**
   - Referral for physical therapy to aid in improving range of motion and strength.

4. **Follow-up Recommendations:**
   - Follow-up visit in 2 weeks for re-evaluation of the knee.
   - Referral to Dr. Emily Carter, a rheumatologist, for further management of rheumatoid arthritis.

5. **Patient Education and Counseling:**
   - Discussed the importance of reducing salt intake to help manage edema.
   - Encouraged the patient to elevate the legs when possible to reduce swelling.
   - Advised the patient to continue using compression socks as a preventive measure.
   - Discussed the potential need for surgical intervention if conservative measures fail to improve symptoms.

6. **Medication Adjustment:**
   - Continue Xeljanz 11mg daily.
   - Continue Celebrex 200mg daily.
   - Regular monitoring of Xeljanz trough levels every 2 months due to the high-risk medication profile.

**Patient Agreements:**
The patient understands and agrees with the recommended medical treatment plan.

**INSTRUCTIONS:**
The patient will follow up in 2 weeks for re-evaluation and will have an X-ray of the right knee performed. He will also make an appointment with Dr. Emily Carter, a rheumatologist, and ensure regular blood work for Xeljanz trough levels.

\\
\bottomrule
\end{tabular}
\caption{Note Generated by the Model Trained Solely on MedSynth}
\label{fig:MedSynth_note_example}
\end{figure}

\clearpage

\section{Survey on Important Quality Factors}
\label{sec:appendix_5}
\begin{figure}[!ht]
  \includegraphics[width=\columnwidth]{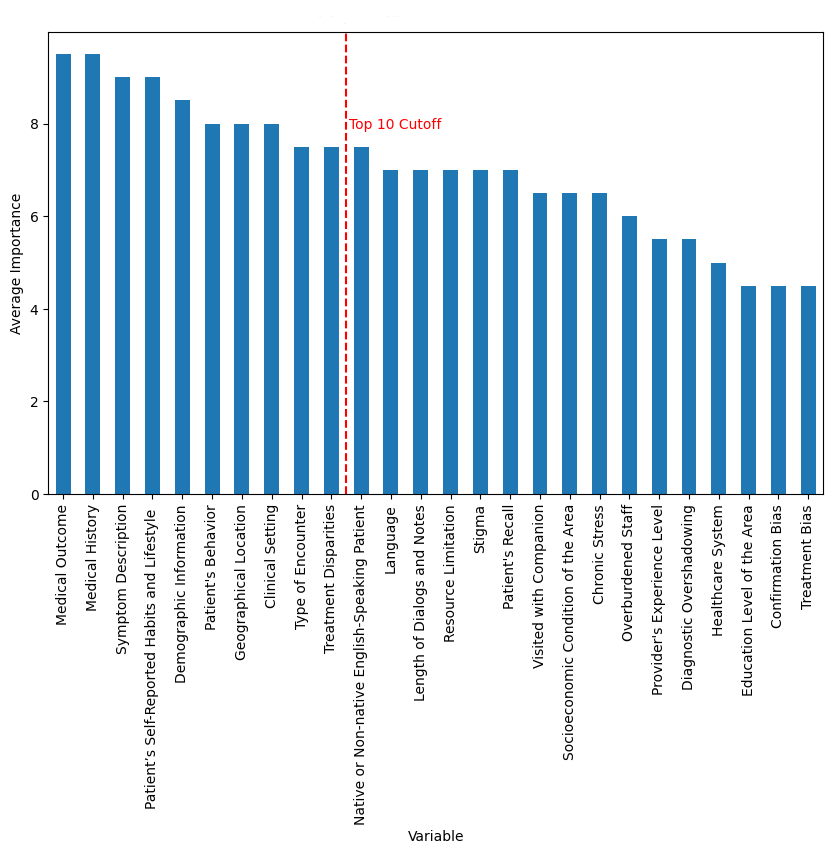}
  \caption{Results of the medical notes quality factors}
  \label{fig:survey_important_variables}
\end{figure}

Figure \ref{fig:survey_important_variables} shows the results of our survey on the important variables in determining the quality of medical notes.

We calculated Intraclass Correlation Coefficient (ICC) \cite{koo2016guideline} to measure the agreement between the annotators. There are multiple forms of ICC. Given the characteristics of our survey and the guide from \cite{koo2016guideline}, we used \( ICC(2, k) \). The observed \( ICC(2, k) \) score was \(  0.56 \), which is considered a moderate score.

\newpage

\section{Sample of Notes and Feedback in Data Generation Pipeline}
\label{sec:appendix_6}

\begin{figure}[hbt!]
\centering
\begin{tabular}{p{0.48\textwidth}} % adjust width to fit half-page
\toprule
\tiny % sets the font size to tiny, the smallest available

DECISION: NoGo

Reasons:

1. Condition (a) is not met: The scenario does not differ sufficiently from the previously approved scenario. The variables such as Medical Outcome, Symptom Description, Patient's Behavior, Geographical Location, Clinical Setting, and Type of Encounter are very similar to the previously approved scenario. There are not enough differences in at least 4 out of the 13 variables.

Feedback:

- Consider varying the demographic information, such as age, gender, or ethnicity, to create a more distinct scenario.

- Introduce a different medical history or chronic condition to differentiate the patient's background.

- Alter the geographical location or clinical setting to provide a new context for the scenario.

- Modify the patient's self-reported habits and lifestyle to reflect a different lifestyle or behavior pattern.

- Explore different types of encounters or treatment disparities to add variety to the scenario.
\\
\bottomrule
\end{tabular}
\caption{An example of feedback provided from Scenario Judge Agent to Scenario Provider Agent}
\label{fig:MedSynth_feedback_example}
\end{figure}

\begin{figure}[hbt!]
\centering
\begin{tabular}{p{0.48\textwidth}} % adjust width to fit half-page
\toprule
\tiny % sets the font size to tiny, the smallest available
 **1. Subjective:**  
 
 **Chief Complaint (CC):** 
 
 Moderate wheezing and shortness of breath.  
 
 **History of Present Illness (HPI):** The patient, Maria Gonzalez, is a 35-year-old Hispanic female with a known history of asthma who presents with worsening symptoms over the past three days. She reports moderate wheezing, shortness of breath, chest tightness, and occasional coughing. Symptoms are exacerbated at night and with physical exertion, significantly impacting her daily activities and sleep.  
 
 **Review of Systems (ROS):** 
 
 - **Constitutional:** Reports interrupted sleep averaging 5-6 hours per night due to asthma symptoms. 
 
 - **Respiratory:** Positive for wheezing, shortness of breath, chest tightness, and occasional coughing. 
 
 - **Cardiovascular:** Denies chest pain or palpitations. 
 
 - **Gastrointestinal:** Denies nausea, vomiting, or changes in bowel habits. 
 
 - **Musculoskeletal:** Denies joint pain or swelling. 
 
 - **Neurological:** Denies headaches, dizziness, or syncope.  **2. Objective:**  **Vital Signs:** 
 
 - Temperature: 98.6°F - Pulse: 88 bpm - Respiratory Rate: 22 breaths per minute - Blood Pressure: 120/80 mmHg 
 
 - Pulse Oxygenation: 95\% on room air  **Physical Examination:** 
 
 - **General:** Patient appears in moderate respiratory distress, no cyanosis. 
 
 - **Respiratory:** Bilateral wheezing on auscultation, decreased breath sounds in the lower lobes. 
 
 - **Cardiovascular:** Regular rate and rhythm; no murmurs, gallops, or rubs. 
 
 - **Abdomen:** Soft, non-tender, no hepatosplenomegaly. 
 
 - **Extremities:** No clubbing, cyanosis, or edema.

 **3. Assessment:**  
 
 **Primary Diagnosis:** Mild persistent asthma with (acute) exacerbation.
 
 **Differential Diagnoses:** 
 
 - Chronic obstructive pulmonary disease (COPD) 
 
 - Respiratory tract infection 
 
 - Cardiac-related dyspnea

 **4. Plan:**  
 
 **Medical Treatment:** 
 
 - **Albuterol Inhaler 90 mcg:** 2 puffs every 4-6 hours as needed for wheezing or shortness of breath. 
 
 - **Fluticasone Propionate Inhaler 110 mcg:** 2 puffs twice daily.  **Follow-up and Referrals:** 
 
 - Schedule follow-up appointment in 2 weeks. 
 
 - Referral to Pulmonologist, Dr. Amanda Smith, for further asthma management.  
 
 **Investigations:** 
 
 - Spirometry to evaluate lung function.
 
 - Complete Blood Count (CBC) and Allergy Panel to evaluate potential allergens and eosinophilia.
 
 **Lifestyle and Patient Education:** 

- Encourage avoidance of known allergens. 

- Emphasize the importance of regular medication adherence. 

- Discuss and implement an asthma action plan. 

- Instruct on proper inhaler technique.  

**Patient Agreements:** 

The patient understands and agrees with the recommended medical treatment plan and follow-up. She is highly cooperative and adherent to medical advice.
\\
\bottomrule
\end{tabular}
\caption{An example of SOAP note generated by the Note Generation pipeline}
\label{fig:MedSynth_note_examples}
\end{figure}

\end{document}